\documentclass{clv3}

\usepackage{adjustbox}
\usepackage{amsmath}
\usepackage{hyperref}
\usepackage{lmodern} 
\usepackage{slantsc} 
\usepackage{tabularx}
\usepackage{xcolor}
\usepackage{tikz}
\usepackage{lscape}
\usepackage{multirow}
\usepackage{pdflscape}
\usepackage{threeparttable}

\newcommand*{\edit}[2]{[\textit{#1} $\rightarrow$ \textit{#2}]}

\makeatletter
\newcommand{\leqnos}{\tagsleft@true\let\veqno\@@leqno}
\newcommand{\reqnos}{\tagsleft@false\let\veqno\@@eqno}
\reqnos
\makeatother

\usepackage{pifont}
\newcommand{\cmark}{\ding{51}}%
\newcommand{\xmark}{\ding{55}}%

\newcolumntype{L}[1]{>{\raggedright\arraybackslash}p{#1}}
\newcolumntype{C}[1]{>{\centering\arraybackslash}p{#1}}
\newcolumntype{R}[1]{>{\raggedleft\arraybackslash}p{#1}}

\DeclareMathOperator*{\argmax}{arg\,max}

\begin{document}

\issue{x}{y}{2023}

\runningtitle{Grammatical Error Correction: A Survey}
\runningauthor{Bryant, Yuan, Qorib, Cao, Ng and Briscoe}

\pageonefooter{Submission received: 08 November 2022; revised version received: 23 March 2023; accepted for publication: 05 April 2023.}

\title{Grammatical Error Correction: \\A Survey of the State of the Art}

\author{Christopher Bryant\thanks{Most of the work in this paper was undertaken while these authors were supported by the ALTA Institute at the University of Cambridge.}}
\affil{ALTA Institute, University of Cambridge \\\url{cjb255@cam.ac.uk} \\Writer, Inc.\\\url{christopher@writer.com}}

\author{Zheng Yuan*}
\affil{Department of Informatics, \\King's College London\\\url{zheng.yuan@kcl.ac.uk}}

\author{Muhammad Reza Qorib}
\affil{National University of Singapore\\\url{mrqorib@u.nus.edu}}

\author{Hannan Cao}
\affil{National University of Singapore\\\url{caoh@u.nus.edu}}

\author{Hwee Tou Ng}
\affil{National University of Singapore\\\url{nght@comp.nus.edu.sg}}

\author{Ted Briscoe*}
\affil{Mohamed bin Zayed University of Artificial Intelligence\\\url{ted.briscoe@mbzuai.ac.ae}}

\maketitle

\begin{abstract}
Grammatical Error Correction (GEC) is the task of automatically detecting and correcting errors in text. The task not only includes the correction of grammatical errors, such as missing prepositions and mismatched subject-verb agreement, but also orthographic and semantic errors, such as misspellings and word choice errors respectively. The field has seen significant progress in the last decade, motivated in part by a series of five shared tasks, which drove the development of rule-based methods, statistical classifiers, statistical machine translation, and finally neural machine translation systems which represent the current dominant state of the art. In this survey paper, we condense the field into a single article and first outline some of the linguistic challenges of the task, introduce the most popular datasets that are available to researchers (for both English and other languages), and summarise the various methods and techniques that have been developed with a particular focus on artificial error generation. We next describe the many different approaches to evaluation as well as concerns surrounding metric reliability, especially in relation to subjective human judgements, before concluding with an overview of recent progress and suggestions for future work and remaining challenges. We hope that this survey will serve as comprehensive resource for researchers who are new to the field or who want to be kept apprised of recent developments. 
\end{abstract}

\section{Introduction}
\label{sec:intro}




Writing is a learnt skill that is particularly challenging for
non-native language users. We all make occasional mistakes with
punctuation, spelling and minor infelicities of word choice in our
native language, but non-native writers often also struggle to create
grammatical and comprehensible texts. Research in the field of Natural
Language Processing (NLP) has addressed the problem of `ill-formed input' at
least since the 1980s because downstream parsing of text usually
collapsed unless input was grammatical \citep{kwasny-sondheimer-1981-relaxation,jensen-etal-1983-parse}. However, useful
applications able to significantly assist non-native writers only
began to appear in the 2000s, such as ETS's Criterion \citep{burstein2003criterion} and Microsoft's ESL Assistant \citep{leacock-etal-2009-user}. These systems were largely based on hand-coded `mal-rules' applied to the output from robust parsers which suggested corrections for errors.

Around the same time, researchers began exploring more data-driven
approaches using supervised machine learning models built from
annotated corpora of errorful text with exemplary corrections \citep{brockett-etal-2006-correcting,de-felice-pulman-2008-classifier,rozovskaya-roth-2010-training,tetreault-etal-2010-using,dahlmeier-ng-2011-grammatical}.
The Helping Our Own (HOO) shared task \citep{hoo2012}, which
attracted 14 research groups to compete and report their results on correcting English determiner and preposition choice errors using
the First Certificate in English (FCE) corpus \citep{yannakoudakis2011}, marked with
hindsight the turning point from rule-based to data-driven methods as well
as burgeoning interest in the task. \citet{leacock2014automated} subsequently published a
book length survey summarising progress in the field up to this point.

The next decade has seen three further expanded shared tasks and an
explosion of research and publications, both from participants in
these competitions and others benchmarking their systems against the
released test sets. Performance has increased roughly three-fold, and
today, most state-of-the-art systems treat the task as one of
`translation' from errorful to corrected text, including the latest
system deployed in Google Docs and Gmail \citep{hoskere_2019}. Recently,
\citet{wang2020comprehensive} provided another detailed survey of work on
grammatical error correction summarising most work published since
\citet{leacock2014automated}. In this article, we provide a more in-depth focus on very recent deep learning based approaches to the task as
well as a more detailed discussion of the nature of the task, its
evaluation, and other remaining challenges (such as multilingual GEC) in order to better equip
researchers with the insights required to be able to contribute to further progress. 

\subsection{The Task}
\label{sec:task}

The definition of a grammatical error is surprisingly difficult. Some
types of spelling errors (such as {\it accomodation} with a single
{\it m}) are about equally distributed between native and non-native
writers and have no grammatical reflexes, so could be reasonably
excluded. Others, such as {\it he eated}, are boundary cases as they
result from over-regularisation of morphology, whilst {\it he would
  eated} is clearly ungrammatical in the context of a modal auxiliary
verb. At the interpretative boundary, infelicitous discourse
organisation, such as {\it Kim fell. Sandy pushed him.} where the
intention is to explain why Kim fell, is not obviously a grammatical
error {\it per se} but nevertheless can be `corrected' via a tense
change ({\it Sandy had pushed him.}) as opposed to a reordering of the
sentences. Other tense changes which can span sentences appear more
grammatical, such as {\it Kim will make Sandy a sandwich. Sandy ate
  it.}, as the discourse is incoherent and correction will require a
tense change in one or other sentence.

In practice, the task has increasingly been defined in terms of what
corrections are annotated in corpora used for the shared tasks. These
use a variety of annotation schemes but all tend to adopt minimal
modifications of errorful texts to create error-free text with the
same perceived meaning. Other sources of annotated data, such as that
sourced from the online language learning platform Lang-8 \citep{lang8,tajiri-etal-2012-tense}, often contain much more
extensive rewrites of entire paragraphs of text. Given this
resource-derived definition of the task, systems are evaluated on
their ability to correct all kinds of mistakes in text, including spelling and discourse level errors that have no or little grammatical reflex. The term `Grammatical' Error Correction is thus something of a misnomer, but is nevertheless now commonly understood to encompass errors that are not always strictly grammatical in nature. A more descriptive term is {\it Language Error Correction}.

\autoref{tab:gec_examples} provides a small sample of (constructed) examples that
illustrate the range of errors to be corrected and some of the issues
that arise with the precise definition and evaluation of the task.
\begin{table}
\centering
\begin{tabular}{lrr}
\textbf{Type} & \textbf{Error} & \textbf{Correction} \\   
\hline
Preposition  & I sat in the talk           & I sat in on the talk \\ 
Morphology   & dreamed                     & dreamt \\ 
Determiner   & I like the ice cream        & I like ice cream \\ 
Tense/Aspect & I like kiss you             & I like kissing you \\
Agreement    & She likes him and kiss him  & She likes him and kisses him \\ 
Syntax       & I have not the book         & I do not have the book \\
Punctuation  & We met they talked and left & We met, they talked and left \\
Unidiomatic  & We had a big conversation   & We had a long conversation \\
Multiple     & I sea the see from the seasoar & I saw the sea from the seesaw\\
\end{tabular}
\caption{Example error types}
\label{tab:gec_examples}
\end{table}
Errors can be classified into three broad categories: \textbf{replacement} errors, such as {\it dreamed} for {\it
  dreamt} in the second example; \textbf{omission} errors, such as {\it on}
in the first example; and \textbf{insertion} errors, such as {\it the} in
the third example. Some errors are complex in the sense that their
correction requires a sequence of replacement, omission or insertion
steps to correct, as with the syntax example. Sentences may also
contain multiple distinct errors that require a sequence of
corrections, as in the multiple example. Both the classification and
specification of correction steps for errors can be and has been
achieved using different schemes and approaches. For instance,
correction of the syntax example involves transposing two adjacent words
so we could introduce a fourth broad class and correction step of
transposition (word order). All extant annotation schemes break these broad classes
down into further subclasses based on the part-of-speech of the words
involved, and perceived morphological, lexical, syntactic, semantic or
pragmatic source of the error. The schemes vary in the number of such
distinctions, ranging from just over two dozen (NUCLE: \citep{nucle}) to almost a hundred
(CLC: \citep{clc}). The schemes also identify different error spans in source
sentences and thus suggest different sets of edit operations to obtain
the suggested corrections. For instance, the agreement error example
might be annotated as \textit{She likes him and} \edit{kiss}{kisses} 
\textit{him} at the token level or simply \edit{$\epsilon$}{es} at the character level. These differing
annotation decisions affected the evaluation of system performance in
artefactual ways, so a two-stage automatic standardisation process was developed, ERRANT \citep{felice-etal-2016-automatic,bryant-etal-2017-automatic}, which maps parallel errorful and corrected sentence pairs to a single annotation scheme using a linguistically-enhanced alignment algorithm and series of error type classification rules. This scheme uses 25 main error type categories, based
primarily on part-of-speech and morphology, which are further subdivided into missing (omission), unnecessary (insertion) and replacement errors. This
approach allows consistent automated training and evaluation of
systems on any or all parallel corpora as well as supporting a
more fined-grained analysis of the strengths and weaknesses of
systems in terms of different error types.

Ultimately however, the correction of errors requires an understanding of the
communicative intention of the writer. For instance, the determiner
example in \autoref{tab:gec_examples} implicitly assumes a `neutral' context where the
intent is to make a statement about generic ice-cream rather than a
specific instance. In a context where, say, a specific ice-cream
dessert is being compared to an alternative dessert, then the
determiner is felicitous. Similarly the preposition omission error
might not be an error if the writer is describing a context in which a
talk was oversubscribed and many attendees had to stand because of a
lack of seats. Though annotators will most likely take both the context and perceived writer's intention into account when identifying errors, GEC itself is instead often framed as an isolated sentence-based task that ignores the wider context. This can introduce noise in the task in that errorful sequences in context may appear correct in isolation out of context. A related issue is that correction may not only depend on communicative intent, but also factors such as dialect and
genre. For example, correcting {\it dreamed} to {\it dreamt} may be
appropriate if the target is British English, but incorrect for
American English.

A larger issue arises with differing possibilities for correction. For
example, correcting the tense/aspect example to {\it kissing} or {\it
  to kiss} in the context of {\it likes} seems equally correct.
However, few existing corpora provide more than one possibility which
means the true performance of systems is often underestimated.
However, the same two corrections are not equally correct as
complements of a verb such as {\it try} depending on whether the
context implies that a kissing event occurred or not. The issue of
multiple possible corrections arises with many if not most examples:
for instance {\it I haven't the book}, {\it We met them, talked and
  left}, {\it We had an important conversation}, {\it The sea I see
  from the seesaw (is calm)} are all plausible alternative corrections
for some of the examples in \autoref{tab:gec_examples}. For this reason, several of the shared
tasks have also evaluated performance on grammatical error detection,
as this is valuable in some applications. Recently, some work has explored treating the GEC
task as one of document-level correction (e.g. \citet{chollampatt-etal-2019-cross,yuan-bryant-2021-document})
which, in principle, could ameliorate some of these issues but is
currently hampered by a lack of appropriately structured corpora.

\subsection{Survey Structure}

We organise the remainder of this survey according to \autoref{tab:summary}. We note that our taxonomy of core approaches, additional techniques and data augmentation (\autoref{sec:core_app}-\ref{sec:aeg}) is similar to that of \citet{wang2020comprehensive}, because these sections contain unavoidable discussions of well-established techniques. We nevertheless believe this is the most effective way of categorising this information and have endeavoured to make the sections complementary in terms of the insights and information they provide. 

\begin{table}[t!]
\centering
\footnotesize
\begin{tabularx}{\textwidth}{llX}
      & \textbf{Subject} & \textbf{Topics} \\
\hline
\autoref{sec:data} & Data & Data collection and annotation, benchmark English datasets, other English datasets, non-English datasets \\
\autoref{sec:core_app} & Core Approaches & Classifiers, statistical machine translation, neural machine translation, edit-based approaches, language models and low-resource systems \\
\autoref{sec:add_tech} & Additional Techniques & Reranking, ensembling and system combination, multi-task learning, custom inference methods, contextual GEC, Generative Adversarial Networks (GANs) \\
\autoref{sec:aeg} & Data Augmentation & Rule-based noise injection, probabilistic error patterns, back-translation, round-trip translation  \\
\autoref{sec:eval} & Evaluation & Benchmark metrics, reference-based metrics, reference-less metrics, metric reliability and human judgements, common experimental settings \\
\autoref{sec:sys_comp} & System Comparison & Recent state-of-the-art systems \\
\autoref{sec:chal} & Future Challenges & Domain generalisation, personalised systems, feedback comment generation, model interpretability, semantic errors, contextual GEC, system combination, training data selection, unsupervised approaches, multilingual GEC, spoken GEC, improved evaluation \\
\autoref{sec:conc} & Conclusion & - \\
\hline
\end{tabularx}%
\caption{Survey structure}
\label{tab:summary}
\end{table}
\section{Data}
\label{sec:data}

Like most tasks in NLP, the cornerstone of modern GEC systems is data. State-of-the-art neural models depend on millions or billions of words and the quality of this data is paramount to model success. Collecting high quality annotated data is a slow and laborious process however, and there are fewer resources available in GEC than other fields such as machine translation. This section hence first outlines some key considerations of data collection in GEC and highlights the importance of robust annotation guidelines. It next introduces the most commonly used corpora in English, as well as some less commonly used corpora, before concluding with GEC corpora for other languages. Artificial data has also become a popular topic in recent years, but this section only covers human annotated data; artificial data will be covered in \autoref{sec:aeg}.

\subsection{Annotation Challenges}
\label{sec:ant_chal}

As mentioned in \autoref{sec:task}, the notion of a grammatical error is hard to define as different errors may have different scope (e.g. local vs. contextual), complexity (e.g. orthographic vs. semantic) and corrections (e.g. \edit{this books}{this book} vs. \edit{this books}{these books}. Human annotation is thus an extremely cognitively demanding task and so clear annotation guidelines are a crucial component of dataset quality. This section briefly outlines three important aspects of data collection: Minimal vs. Fluent Corrections, Annotation Consistency, and Preprocessing Challenges.

\paragraph{Minimal vs. Fluent Corrections} Most GEC corpora have been annotated on the principle of \textit{minimal corrections}, i.e. annotators should make the minimum number of changes to make a text grammatical. \citet{sakaguchi-etal-2016-reassessing} argue, however, that this can often lead to corrections that sound unnatural, and so it would be better to annotate corpora on the principle of \textit{fluent corrections} instead. Consider the following example: 

\begin{center}
\begin{tabular}{l|l}
Original & I want explain to you some interesting part from my experience. \\
Minimal  & I want \underline{to} explain to you some interesting \underline{parts of} my experience. \\
Fluent   & I want \underline{to tell you about} some interesting \underline{parts of} my experience. \\
\end{tabular}  
\end{center}

\noindent While the minimal correction primarily inserts a missing infinitival \textit{to} before \textit{explain} to make the sentence grammatical, the fluent correction also changes \textit{explain} to \textit{tell you about} because it is more idiomatic to tell someone about an experience rather than explain an experience.

One of the main challenges of this distinction, however, is that it is very difficult to draw a line between what constitutes a minimal correction and what constitutes a fluent correction. This is because minimal corrections (e.g. missing determiners) are a subset of fluent corrections, and so there cannot be fluent corrections without minimal corrections. It is also the case that minimal corrections are typically easier to make than fluent corrections (for both humans and machines), although it is undeniable that fluent corrections are the more desirable outcome. Ultimately, although it is very difficult to precisely define a fluent correction, annotation guidelines should nevertheless attempt to make clear the extent to which annotators are expected to edit. 

\paragraph{Annotation Consistency} A significant challenge of human annotation is that corrections are subjective and there is often more than one way to correct a sentence \citep{bryant2015,choshen-abend-2018-inherent}. It is nevertheless important that annotators attempt to be consistent in their judgements, especially if they are explicitly annotating edit spans. For example the edit \edit{has eating}{was eaten} can also be represented as \edit{has}{was} and \edit{eating}{eaten}, and this choice not only affects data exploration and analysis, but can also have an impact on edit-based evaluation. Similarly, the edit \edit{the informations}{information} can also be represented as \edit{the}{$\epsilon$} and \edit{informations}{information}, but the latter may be more intuitive because it represents two independent edits of clearly distinct types. Explicit error type classification is thus another important aspect of annotator consistency, as an error type framework (if any) not only increases the cognitive burden on the annotator, but also might influence an annotator towards a particular correction given the error types that are available \citep{sakaguchi-etal-2016-reassessing}. Ultimately, if annotators are tasked with explicitly defining the edits they make to correct a sentence, annotator guidelines must clearly define the notion of an edit. 

\paragraph{Preprocessing Challenges} While human annotators are trained to correct natural text, GEC systems are typically trained to correct word tokenised sentences (mainly for evaluation purposes). This mismatch means human annotations typically undergo several preprocessing steps in order to produce the desired output format \citep{bryant-2016-issues}. The first of these transformations involves converting character-level edits to token-level edits. While this is often straightforward, it can sometimes be the case that a human-annotated character span does not map to a complete token; e.g. \edit{ing}{ed} to denote the edit \edit{dancing}{danced}. Although such cases can often (but not always) be resolved automatically, e.g., by expanding the character spans of the edit or calculating token alignment, they can also be reduced by training annotators to explicitly annotate longer spans rather than sub-words. 

The second transformation involves sentence tokenisation, which is potentially more complex given human edits may change sentence boundaries; e.g. \edit{A. B, C.}{A, B. C.}. Sentences are nevertheless typically tokenised based solely on the original text, with the acknowledgement that some may be sentence fragments (to be joined with the following sentence) and that edits which cross sentence boundaries are ignored (e.g. \edit{. Because}{, because}. It is worth noting that this issue only affects sentence-based GEC systems (the vast majority) but paragraph or document-based systems are unaffected. 

\subsection{English Datasets}

A small number of English GEC datasets have become popular for training and testing GEC systems, mostly as a result of shared tasks.\footnote{\url{https://www.cl.cam.ac.uk/research/nl/bea2019st/\#data}} This section introduces them as well as other less popular datasets for English (\autoref{tab:datasets}). We acknowledge that this is by no means an exhaustive list, but highlight datasets that have gained some traction in the last few years. 

\subsubsection{Benchmark English Datasets}

\begin{table}[t]
  \centering
  \footnotesize
\begin{adjustbox}{width=\textwidth}
\begin{tabular}{L{1.7cm}|l|cc|r|c|R{0.8cm}|R{1.35cm}|L{1.95cm}}
\textbf{Corpus} & \textbf{Use} & \multicolumn{1}{r|}{\textbf{Sents}} & \multicolumn{1}{r|}{\textbf{Toks}} & \textbf{Refs} & \multicolumn{1}{L{0.8cm}|}{\textbf{Edit Spans}} & \textbf{Error Types} & \textbf{Level} & \textbf{Domain} \\
\hline\hline
FCE   & Train & \multicolumn{1}{r|}{28.3k} & \multicolumn{1}{r|}{454k} & 1     & \cmark & 71    & B1-B2 & Exams \\
      & Dev   & \multicolumn{1}{r|}{2.2k} & \multicolumn{1}{r|}{34.7k} & 1     & \cmark & 71    & B1-B2 & Exams \\
      & Test  & \multicolumn{1}{r|}{2.7k} & \multicolumn{1}{r|}{41.9k} & 1     & \cmark & 71    & B1-B2 & Exams \\
\hline
NUCLE & Train & \multicolumn{1}{r|}{57.1k} & \multicolumn{1}{r|}{1.16m} & 1     & \cmark & 28    & C1    & Essays \\
\hline
CoNLL-2013 & Dev/Test   & \multicolumn{1}{r|}{1.4k} & \multicolumn{1}{r|}{29.2k} & 1     & \cmark & 28    & C1    & Essays \\
\hline
CoNLL-2014 & Test  & \multicolumn{1}{r|}{1.3k} & \multicolumn{1}{r|}{30.1k} & 2-18 & \cmark & 28    & C1    & Essays \\
\hline
Lang-8 & Train & \multicolumn{1}{r|}{1.03m} & \multicolumn{1}{r|}{11.8m} & 1-8 & \xmark & 0     & A1-C2? & Web \\
\hline
JFLEG & Dev   & \multicolumn{1}{r|}{754} & \multicolumn{1}{r|}{14.0k} & 4     & \xmark & 0     & A1-C2? & Exams \\
      & Test  & \multicolumn{1}{r|}{747} & \multicolumn{1}{r|}{14.1k} & 4     & \xmark & 0     & A1-C2? & Exams \\
\hline
W\&I+ & Train & \multicolumn{1}{r|}{34.3k} & \multicolumn{1}{r|}{628k} & 1     & \cmark & 55    & A1-C2 & Exams \\
LOCNESS    & Dev   & \multicolumn{1}{r|}{4.4k} & \multicolumn{1}{r|}{87.0k} & 1     & \cmark & 55    & A1-Native & Exams, Essays \\
(BEA-2019) & Test  & \multicolumn{1}{r|}{4.5k} & \multicolumn{1}{r|}{85.7k} & 5     & \cmark & 55    & A1-Native & Exams, Essays \\
\hline\hline
CLC   & Train & \multicolumn{1}{r|}{1.96m} & \multicolumn{1}{r|}{29.1m} & 1     & \cmark & 77    & A1-C2 & Exams \\
\hline
EFCamDat & Train & \multicolumn{1}{r|}{4.60m} & \multicolumn{1}{r|}{56.8m} & 1     & \cmark & 25    & A1-C2 & Exams \\
\hline
WikEd & Train & \multicolumn{1}{r|}{28.5m} & \multicolumn{1}{r|}{626m} & 1     & \xmark & 0     & Native & Wiki \\
\hline
AESW  & Train & \multicolumn{1}{r|}{1.20m} & \multicolumn{1}{r|}{28.4m} & 1     & \cmark & 0     & C1-Native & Science \\
      & Dev   & \multicolumn{1}{r|}{148k} & \multicolumn{1}{r|}{3.51m} & 1     & \cmark & 0     & C1-Native & Science \\
      & Test  & \multicolumn{1}{r|}{144k} & \multicolumn{1}{r|}{3.45m} & 1     & \cmark & 0     & C1-Native & Science \\
\hline
GMEG  & Dev   & \multicolumn{1}{r|}{2.9k} & \multicolumn{1}{r|}{60.9k} & 4     & \xmark & 0     & B1-B2, Native & Exams, Web, Wiki \\
      & Test  & \multicolumn{1}{r|}{2.9k} & \multicolumn{1}{r|}{61.5k} & 4     & \xmark & 0     & B1-B2, Native & Exams, Web, Wiki \\
\hline
CWEB  & Dev   & \multicolumn{1}{r|}{6.7k} & \multicolumn{1}{r|}{148k} & 2     & \cmark & 55    & Native & Web \\
      & Test  & \multicolumn{1}{r|}{6.8k} & \multicolumn{1}{r|}{149k} & 2     & \cmark & 55    & Native & Web \\
\hline
GHTC  & Train? & \multicolumn{2}{c|}{353k edits only} & 1     & \cmark & 0     & Native? & Documentation \\
\hline\hline
\end{tabular}%
\end{adjustbox}
\caption{Human-annotated GEC datasets for English. The top half are commonly used to benchmark GEC systems. A question mark (?) indicates unknown or approximated information. CEFR levels: beginner (A1-A2), intermediate (B1-B2), advanced (C1-C2).}
  \label{tab:datasets}%
\end{table}%

\paragraph{FCE} The First Certificate in English (FCE) corpus \citep{yannakoudakis2011} is a public subset of the Cambridge Learner Corpus (CLC) \citep{clc} that consists of 1,244 scripts ($\sim$531k words) written by international learners of English as a second language (L2 learners). Each script typically contains two answers to a prompt in the style of a short essay, letter, or description, and each answer has been corrected by a single annotator who has identified and classified each edit according to a framework of 88 error types \citep{clc} (71 unique error types are represented in the FCE). The authors are all intermediate level (B1-B2 level on the Common European Framework of Reference for Languages (CEFR) \citep{cefr}) and the data is split into a standard training, development and test set. The FCE was used as the official dataset of the HOO-2012 shared task \citep{hoo2012}, one of the official training datasets of the BEA-2019 shared task \citep{bea2019}, and has otherwise commonly been used for grammatical error detection \citep{rei-yannakoudakis-2016,bell-etal-2019-context,yuan-etal-2021-multi}. It also contains essay level scores, as well as other limited metadata about the learner, and has been used for automatic essay scoring (AES) (e.g. \citet{ijcai2019-879}). 

\paragraph{NUCLE/CoNLL} The National University of Singapore Corpus of Learner English \mbox{(NUCLE)}  \citep{nucle} consists of 1,397 argumentative essays ($\sim$1.16m words) written by NUS undergraduate students who needed L2 English language support. The essays, which are approximately C1 level, are written on a diverse range of topics including technology, healthcare, and finance, and were each corrected by a single annotator who identified and classified each edit according to a framework of 28 error types. NUCLE was used as the official training corpus of the CoNLL-2013 and CoNLL-2014 shared tasks \citep{conll2013,conll2014} as well as one of the official training datasets of the BEA-2019 shared task \citep{bea2019}. The CoNLL-2013 and CoNLL-2014 test sets were annotated under similar conditions to NUCLE and respectively consist of 50 essays each ($\sim$30k words) on the topics of i) surveillance technology and population aging, and ii) genetic testing and social media. The CoNLL-2014 test set was also doubly annotated by 2 independent annotators, resulting in 2 sets of official reference annotations; \citet{bryant2015} and \citet{sakaguchi-etal-2016-reassessing} subsequently collected another 8 sets of annotations each for a total of 18 sets of reference annotations. The CoNLL-2013 dataset is now occasionally used as a development set, while the CoNLL-2014 dataset is one of the most commonly used benchmark test sets. One limitation of the CoNLL-2014 test set is that it is not very diverse given that it consists entirely of essays written by a narrow range of learners on only two different topics.

\paragraph{Lang-8} The Lang-8 Corpus of Learner English \citep{lang8,tajiri-etal-2012-tense} is a preprocessed subset of the multilingual Lang-8 Learner Corpus \citep{mizumoto-etal-2011-mining}, which consists of 100,000 submissions ($\sim$11.8m words) to the language learning social network service, Lang-8.\footnote{\url{http://lang-8.com}} The texts are wholly unconstrained by topic, and hence include the full range of ability levels (A1-C2), and were written by international L2 English language learners with a bias towards Japanese L1 speakers. Although Lang-8 is one of the largest publicly available corpora, it is also one of the noisiest as corrections are provided by other users rather than professional annotators. A small number of submissions also contain multiple sets of corrections, but all annotations are provided as parallel text and so do not contain explicit edits or error types. Lang-8 was also one of the official training datasets of the BEA-2019 shared task \citep{bea2019}.

\paragraph{JFLEG} The Johns Hopkins Fluency-Extended GUG corpus (JFLEG)  \citep{jfleg} is a collection of 1,501 sentences ($\sim$28.1k words) split roughly equally into a development and test set. The sentences were randomly sampled from essays written by L2 learners of English of an unspecified ability level \citep{Heilman-et-al-2014} and corrected by crowdsourced annotators on Amazon Mechanical Turk \citep{crowston-2012-amt}. Each sentence was annotated a total of 4 times, resulting in 4 sets of parallel reference annotations, but edits were not explicitly defined or classified. The main innovation of JFLEG is that sentences were corrected to be fluent rather than minimally grammatical (\autoref{sec:ant_chal}). The main criticisms of JFLEG are that it is much smaller than other test sets, the sentences are presented out of context, and it was not corrected by professional annotators \citep{napoles-etal-2019-enabling}.  

\paragraph{W\&I+LOCNESS} The Write \& Improve (W\&I) and LOCNESS corpus \citep{bea2019} respectively consist of 3,600 essays ($\sim$755k words) written by international learners of all ability levels (A1-C2) and 100 essays ($\sim$46.2k words) written by native British/American English undergraduates. It was released as the official training, development and test corpus of the BEA-2019 shared task and was designed to be more balanced than other corpora such that there are roughly an equal number of sentences at each ability level: Beginner, Intermediate, Advanced, Native. The W\&I essays come from submissions to the Write \& Improve online essay-writing platform\footnote{\url{https://writeandimprove.com/}} \citep{yannakoudakis-2018-developing} and the LOCNESS essays, which only comprise part of the development and test sets, come from the LOCNESS corpus \citep{granger1998}. The training and development set essays were each corrected by a single annotator, while the test set essays were corrected by 5 annotators resulting in 5 sets of parallel reference annotations. Edits were explicitly defined, but not manually classified, so error types were added automatically using the ERRANT framework \citep{bryant-etal-2017-automatic}. The test set references are not currently publicly available, so all evaluation on this dataset is done via the BEA-2019 Codalab competition platform,\footnote{\url{https://www.cl.cam.ac.uk/research/nl/bea2019st/\#instr}} which ensures all systems are evaluated in the same conditions. 

\subsubsection{Other English Datasets}

\paragraph{CLC} The Cambridge Learner Corpus (CLC) \citep{clc} is a proprietary collection of over 130,000 scripts ($\sim$29.1m words) written by international learners of English (130 different first language backgrounds) for different Cambridge exams of all levels (A1-C2) \citep{yuan-etal-2016-candidate,bryant-phdthesis}. It is the superset of the public FCE and annotated in the same way.

\paragraph{EFC\textsl{\textsc{am}}D\textsl{\textsc{at}}} The Education First Cambridge Database (EFC\textsc{am}D\textsc{at}) \citep{efcamdat} consists of 1.18m scripts ($\sim$83.5m words) written by international learners of all ability levels (A1-C2) submitted to the English First online school platform. Approximately 66\% of the scripts ($\sim$56.8m words) have been annotated with explicit edits that have been classified according to a framework of 25 error types \citep{huang-2017-efcamdat}. Since the annotations were made by teachers for the purposes of giving feedback to students rather than for GEC system development, they are not always complete (too many corrections may dishearten the learner). 

\paragraph{WikEd} The Wikipedia Edit Error Corpus (WikEd) \citep{wiked_2014} consists of tens of millions of sentences of revision histories from articles on English Wikipedia. The texts are written and edited by native speakers rather than L2 learners and not all changes are grammatical edits; e.g. information updates. A preprocessed version of the corpus is available\footnote{\url{https://github.com/snukky/wikiedits}} (28.5m sentences, 626m words) which filters and modifies sentences such that they only contain edits similar to those in NUCLE. The corpus also includes tools to facilitate the collection of similar Wiki-based corpora for other languages.

\paragraph{AESW} The Automatic Evaluation of Scientific Writing (AESW) dataset consists of 316k paragraphs ($\sim$35.5m words) extracted from 9,919 published scientific journal articles and split into a training, development and test set for the AESW shared task \citep{Daudaravicius-et-al-2016}. A majority of the paragraphs come from Physics, Mathematics and Engineering journals and were written by advanced or native speakers. The articles were edited by professional language editors who explicitly identified the required edits but did not classify them by error type. Although large, one of the main limitations of the AESW dataset is that the texts come from a very specific domain and many sentences contain placeholder tokens for mathematical notation and reference citations which do not generalise to other domains. 

\paragraph{GMEG} The Grammarly Multi-domain Evaluation for GEC (GMEG) dataset \citep{napoles-etal-2019-enabling} consists of 5,919 sentences ($\sim$122.4k words) split approximately equally across 3 different domains: formal native, informal native, and learner text. Specifically, the formal text is sampled from the WikEd corpus \citep{wiked_2014}, the informal text is sampled from Yahoo Answers, and the learner text is sampled from the FCE \citep{yannakoudakis2011}. The sentences were sampled at the paragraph level (except for WikEd) to include some context and were annotated by 4 professional annotators to produce 4 sets of alternative references. One of the goals of GMEG was to diversify researchers away from purely L2 learner-based corpora.

\paragraph{CWEB} The Corrected Websites (CWEB) dataset \citep{flachs-etal-2020-grammatical} consists of 13.6k sentences (297k words) sampled from random paragraphs on the web in the CommonCrawl dataset.\footnote{\url{https://commoncrawl.org/}} Paragraphs were filtered to reduce noise (e.g. non-English and duplicates) and loosely defined as formal (``sponsored'') and informal (``generic'') based on the domain of the URL. The paragraphs, which are split equally between a development set and a test set, were doubly annotated by 2 professional annotators and edits were extracted and classified automatically using ERRANT \citep{bryant-etal-2017-automatic}. Like GMEG, one of the aims of CWEB was to introduce a dataset that extended beyond learner corpora. 

\paragraph{GHTC} The GitHub Typo Corpus (GHTC) \citep{hagiwara-mita-2020-github} consists of 353k edits from 203k commits to repositories in the GitHub software hosting website.\footnote{\url{https://github.com/}} All the edits were gathered from repositories that met certain conditions (e.g. a permissive license) and from commits that contained the word `typo' in the commit message. The intuition behind the corpus was that developers often make small commits to correct minor spelling/grammatical errors and that these annotations can be used for GEC. The main limitation of GHTC is that the majority of edits are spelling or orthographic errors from a specific domain (i.e. software documentation) and that the context of the edit is not always a full sentence. 

\subsection{Non-English Datasets}

Although most work on GEC has focused on English, corpora for other languages are also slowly being created and publicly released for the purposes of developing GEC models. This section introduces some of the most prominent (\autoref{tab:other_langs}), along with other relevant resources, but is again by no means an exhaustive list. These resources are ultimately helping to pave the way for research into multilingual GEC \citep{naplava-straka-2019-grammatical,katsumata-komachi-2020-stronger,rothe-etal-2021-simple}. 

\begin{table}[t]
  \centering
  \footnotesize
  \begin{adjustbox}{width=\textwidth}
  \begin{threeparttable}
    \begin{tabular}{l|l|l|r|r|r|c|R{0.8cm}|R{1.35cm}|L{2cm}}
    \textbf{Language} & \textbf{Corpus} & \textbf{Use} & \textbf{Sents} & \textbf{Toks} & \textbf{Refs} & \multicolumn{1}{L{0.8cm}|}{\textbf{Edit Spans}} & \textbf{Error Types} & \textbf{Level} & \textbf{Domain} \\
    \hline\hline
    Arabic & QALB-2014 & Train & 19.4k\tnote{*} & 1m    & 1     & \checkmark & 7     & Native & Web \\
          &       & Dev   & 1k\tnote{*}   & 53.8k & 1     & \checkmark & 7     & Native & Web \\
          &       & Test  & 948\tnote{*}  & 51.3k & 1     & \checkmark & 7     & Native & Web \\
    \cline{2-10}
     & QALB-2015 & Train & 310\tnote{*}  & 43.3k & 1     & \checkmark & 7     & A1-C2 & Essays \\
          &       & Dev   & 154\tnote{*}  & 24.7k & 1     & \checkmark & 7     & A1-C2 & Essays \\
          &       & Test  & 158\tnote{*}  & 22.8k & 1     & \checkmark & 7     & A1-C2 & Essays \\
          &       & Test  & 920\tnote{*}  & 48.5k & 1     & \checkmark & 7     & Native & Web \\
    \hline
    Chinese  &  NLPTEA-2020 & Train  & 1.1k\tnote{$\dagger$} & 36.9k\tnote{$\ddagger$} & 1 & \checkmark & 4 & A1-C2 & Exams \\
        &              & Test   & 1.4k\tnote{$\dagger$} & 55.2k\tnote{$\ddagger$} & 1 & \checkmark & 4 & A1-C2 & Exams \\
    \cline{2-10}
            & NLPCC-2018 & Train  & 717k & 14.1m\tnote{$\ddagger$} & 1-21 & \xmark     &  0     & A1-C2? & Web \\
            &                 & Test   & 2k & 61.3k\tnote{$\ddagger$} & 1-2 & \checkmark & 4     & A1-C2? & Essays \\
    \cline{2-10}
            & MuCGEC & Dev  & 1.1k & 50k\tnote{$\ddagger$} & 2.3 & \checkmark     &  19     & A1-C2? & Exams \\
            &  & Test  & 5.9k & 228k\tnote{$\ddagger$} & 2.3 & \checkmark     &  19     & A1-C2? & Essays, Exams, Web \\
    \hline
   Czech & AKCES-GEC & Train & 42.2k & 447k  & 1     & \checkmark & 25    & A1-Native & Essays, Exams \\
          &       & Dev   & 2.5k  & 28.0k & 2     & \checkmark & 25    & A1-Native & Essays, Exams \\
          &       & Test  & 2.7k  & 30.4k & 2     & \checkmark & 25    & A1-Native & Essays, Exams \\
    \cline{2-10}
     & GECCC & Train & 66.6k & 750k  & 1     & \checkmark & 65    & A1-Native & Essays, Exams, Web \\
          &       & Dev   & 8.5k  & 101k  & 1-2   & \checkmark & 65    & A1-Native & Essays, Exams, Web \\
          &       & Test  & 7.9k  & 98.1k & 2     & \checkmark & 65    & A1-Native & Essays, Exams, Web \\
    \hline
    German & Falko-MERLIN & Train & 19.2k & 305k  & 1     & \checkmark & 56    & A1-C2 & Essays, Exams \\
          &       & Dev   & 2.5k  & 39.5k & 1     & \checkmark & 56    & A1-C2 & Essays, Exams \\
          &       & Test  & 2.3k  & 36.6k & 1     & \checkmark & 56    & A1-C2 & Essays, Exams \\
    \hline
    Japanese & TEC-JL & Test  & 1.9k  & 41.5k\tnote{$\ddagger$} & 2     & \xmark & 0     & A1-C2? & Forum \\
    \hline
    Russian & RULEC-GEC & Train & 5k    & 83.4k & 1     & \checkmark & 23    & C1-C2 & Essays \\
          &       & Dev   & 2.5k  & 41.2k & 1     & \checkmark & 23    & C1-C2 & Essays \\
          &       & Test  & 5k    & 81.7k & 1     & \checkmark & 23    & C1-C2 & Essays \\
    \hline
    Ukrainian & UA-GEC & Train & 18.2k & 285k  & 1     & \checkmark & 4     & B1-Native & Essays, Fiction \\
          &       & Test  & 2.5k  & 43.5k & 1     & \checkmark & 4     & B1-Native & Essays, Fiction \\
    \hline\hline
    \end{tabular}%
\begin{tablenotes}\footnotesize
\item[*] The Arabic datasets are split into documents rather than sentences.
\item[$\dagger$] The Chinese NLPTEA datasets are split into paragraphs (1-5 sentences) rather than sentences. 
\item[$\ddagger$] The Chinese and Japanese datasets are split into characters rather than tokens. \\
\end{tablenotes}
\end{threeparttable}
    \end{adjustbox}
  \caption{Human-annotated GEC datasets for non-English languages. A question mark (?) indicates unknown or approximated information. CEFR levels: beginner (A1-A2), intermediate (B1-B2), advanced (C1-C2).}    
  \label{tab:other_langs}%
\end{table}%

\paragraph{Arabic} The Qatar Arabic Language Bank (QALB) project \citep{zaghouani-etal-2014-large} is an initiative that aims to collect large corpora of annotated Arabic for the purposes of Arabic GEC system development. A subset of this corpus was used as the official training, development and test data of the QALB-2014 and QALB-2015 shared tasks on Arabic text correction \citep{mohit-etal-2014-first,rozovskaya-etal-2015-second}. In particular, QALB-2014 released 21.3k documents (1.1m words) of annotated user comments submitted to the Al Jazeera news website by native speakers, while QALB-2015 released 622 documents (90.8k words) of annotated essays written by the full range of Arabic L2 learners (A1-C2) \citep{zaghouani-etal-2015-correction} along with an additional 920 documents (48.5k words) of unreleased Al Jazeera comments. QALB-2015 thus had 2 test sets: one on native Al Jazeera data and one on Arabic L2 learner essays. In all cases, files were provided at the document level (rather than the sentence level) and edits were explicitly identified by trained annotators and classified automatically using a framework of 7 error types.

\paragraph{Chinese} The Test of Chinese as a Foreign Language (TOCFL) corpus \citep{lee-etal-2018-building} and the \textit{Hanyu Shuiping Kaoshi} (HSK: Chinese Proficiency Test) corpus\footnote{\url{http://yuyanziyuan.blcu.edu.cn/en/info/1043/1501.htm}} \citep{zhang2009} respectively consist of 2.8k essays (1m characters) and 11k essays (4m characters) written by the full range of language learners (A1-C2) who took Mandarin Chinese language proficiency exams. Various subsets of these corpora were used as the official training and test sets in the NLPTEA series of shared tasks on Chinese Grammatical Error Diagnosis (i.e. error detection) between 2014-2020 \citep{nlptea2014,rao2020overview}. The most recent of these shared tasks, NLPTEA-2020, released a total of 2.6k paragraphs (92.1k characters, 1-5 sentences each), which were annotated by a single annotator according to a framework of 4 error types: Redundant (R), Missing (M), Word Selection (S) or Word Order (W). 

The NLPCC-2018 shared task \citep{zhao2018overview}, which was the first shared task on full error correction in Mandarin Chinese, released a further 717k training sentences (14.1m characters) which were extracted from a cleaned subset of Lang-8 user submissions \citep{mizumoto-etal-2011-mining}. Like the Lang-8 Corpus of Learner English, the ability level of the authors in this dataset is unknown and corrections were provided by other users. The test data for this shared task came from the PKU Chinese Learner Corpus and consists of 2000 sentences (61.3k characters) written by foreign college students. All test sentences were first annotated by a single annotator, who also classified edits according to the same 4-error-type framework as NLPTEA, and subsequently checked by a second annotator who was allowed to make changes to the annotations if necessary. 

The Multi-Reference Multi-Source Evaluation Dataset for Chinese Grammatical Error Correction (MuCGEC) \citet{zhang2022} is a new corpus that is intended to be a more robust test set for Chinese GEC. It contains a total of 7063 sentences ($\sim$278k characters) sampled approximately equally from the NLPCC-2018 training set (Lang-8), the NLPCC-2018 test set (PKU Chinese Learner Corpus) and the NLPTEA-2018/2020 test sets (HSK Corpus). All sentences were annotated by multiple annotators, but identical references were removed, so we report an average of 2.3 references per sentence (90\% of all sentences have 1-3 references). Edits were also classified according to a scheme of 19 error types, including 5 main error types and 14 minor sub-types. 

\paragraph{Czech} The AKCES-GEC corpus \citep{naplava-straka-2019-grammatical} consists of 47.3k sentences (505k words) written by both learners of Czech as a second language (from both Slavic and non-Slavic backgrounds) and Romani children who speak a Czech ethnolect as a first language. The essays and exam-style scripts come from the Learner Corpus of Czech as a Second Language (CzeSL) \citep{Rosen:2016a} which falls under the larger Czech Language Acquisition Corpora (AKCES) project \citep{Sebesta:2010}. The essays in the training set were annotated once (1 set of annotations) and the essays in the development and test sets were annotated twice (2 sets of annotations), all with explicit edits that were classified according to a framework of 25 error types. 

The Grammar Error Correction Corpus for Czech (GECCC) \citep{naplava-2022-czech} is an extension of AKCES-GEC that includes both formal texts written by native Czech primary and secondary school students as well as informal website discussions on Facebook and Czech news websites, in addition to the texts written by Czech language learners and Romani children. The total corpus consists of 83k sentences (949k words), all of which were manually annotated (or re-annotated in order to preserve annotation style) by 5 experienced annotators who explicitly identified edits. Edits were then classified automatically by a variant of ERRANT \citep{bryant-etal-2017-automatic} for Czech which included a customised tagset of 65 errors types. GECCC is currently one of the largest non-English corpora and is also larger than most popular English benchmarks. 

\paragraph{German} The Falko-MERLIN GEC corpus \citep{boyd2018using} consists of 24k sentences (381k words) written by learners of all ability levels (A1-C2). Approximately half the data comes from the Falko corpus \citep{reznicek2012falko}, which consists of minimally-corrected advanced German learner essays (C1-C2), while the other half comes from the MERLIN corpus \citep{boyd-etal-2014-merlin}, which consists of standardised German language exam scripts from a wide range of ability levels (A1-C1). Edits were not explicitly annotated, but extracted and classified automatically using a variation of ERRANT \citep{bryant-etal-2017-automatic} which was adapted for German and included a customised tagset for German error types.  

\paragraph{Japanese} The TMU Evaluation Corpus for Japanese Learners (TEC-JL) \citep{koyama-etal-2020-construction} consists of 1.9k sentences (41.5k characters) written by language learners of unknown level (A1-C2?) and submitted to the language learning social network service Lang-8. TEC-JL is a subset of the multilingual Lang-8 Learner Corpus \citep{mizumoto-etal-2011-mining} and was doubly annotated by 3 native Japanese university students (2 sets of annotations) to be a more reliable test set than the original Lang-8 Learner Corpus which can be quite noisy. 

\paragraph{Russian} The Russian Learner Corpus of Academic Writing (RULEC) \citep{Alsufieva2012rulec} consists of essays written by L2 university students and heritage Russian speakers in the United States. A subset of this corpus, 12.5k sentences (206k words), was annotated by 2 native speakers of Russian with backgrounds in linguistics and released as the RULEC-GEC corpus \citep{rozovskaya-roth-2019-grammar}. Edits were explicitly annotated and classified according to a framework of 23 error types. Another corpus of annotated Russian errors, the Russian Lang-8 corpus (RU-Lang8) \citep{trinh-rozovskaya-2021-new}, which is a subset of the aforementioned multilingual Lang-8 Learner Corpus \citep{mizumoto-etal-2011-mining}, was also recently announced, however the data has not yet been publicly released. 

\paragraph{Ukrainian} The UA-GEC corpus \citep{syvokon2021uagec} consists of 20.7k sentences (329k words) written by almost 500 authors from a wide variety of backgrounds (mostly technical and humanities) and ability levels (two-thirds native). The texts cover a wide range of topics including short essays (formal, informal, fictional or journalistic) and translated works of literature, and were annotated by two native speakers with degrees in Ukrainian linguistics. Edits were explicitly annotated and classified according to a scheme of 4 error types: Grammar, Spelling, Punctuation or Fluency. 

\section{Core Approaches}
\label{sec:core_app}

This section introduces some of the core approaches to GEC including classifiers (statistical and neural), machine translation (statistical and neural), edit-based approaches and language models. We provide a high level overview of how each of these approaches works and highlight notable models that have led to breakthroughs in system development. These approaches provide the foundation on which additional techniques (\autoref{sec:add_tech}) and artificial error generation (\autoref{sec:aeg}) are built.

\subsection{Classifiers}

Machine learning classifiers were historically one of the most popular approaches to GEC. The main reason for this was that some of the most common error types for English as a second language (ESL) learners, such as article and preposition errors, have small confusion sets and so are well-suited to multi-class classification. For example, it is intuitive to build a classifier that predicts one of \{\textit{a/an}, \textit{the}, \textit{$\epsilon$}\} before every noun phrase in a sentence. To do this, a classifier receives a number of features representing the context of the analysed word or phrase in a sentence and outputs a predicted class that constitutes a correction. Errors are flagged and corrected by comparing the original word used in the text with the most likely candidate predicted by the classifier. This approach has been applied to several common error types including:

\begin{itemize}
    \item articles~\citep{lee-2004-automatic,han_chodorow_leacock_2006,de-felice-2008-automatic,gamon-etal-2008-using,gamon-2010-using,dahlmeier-ng-2011-grammatical,kochmar-etal-2012-hoo,rozovskaya-roth-2013-joint,rozovskaya-roth-2014-building};
    
    \item prepositions~\citep{chodorow-etal-2007-detection,de-felice-2008-automatic,gamon-etal-2008-using,tetreault-chodorow-2008-ups,gamon-2010-using,dahlmeier-ng-2011-grammatical,kochmar-etal-2012-hoo,rozovskaya-roth-2013-joint,rozovskaya-roth-2014-building};
    
    \item noun number~\citep{berend-etal-2013-lfg,van-den-bosch-berck-2013-memory,jia-etal-2013-grammatical,xiang-etal-2013-hybrid,yoshimoto-etal-2013-naist,kunchukuttan-etal-2014-tuning};
    
    \item verb form~\citep{lee-seneff-2008-correcting,tajiri-etal-2012-tense,van-den-bosch-berck-2013-memory,jia-etal-2013-grammatical,rozovskaya-roth-2013-joint,rozovskaya-roth-2014-building,rozovskaya-etal-2014-correcting}.
    
\end{itemize}

Training examples consisting of native and/or learner data are represented as vectors of features that are considered useful for the error type. Since the most useful features often depend on the word class, it is necessary to build separate classifiers for each error type and most of the prior classification-based approaches have focused on feature engineering. For the vast majority of syntactically-motivated errors, features such as contextual word and part-of-speech (POS) n-grams, lemmas, phrase constituency information and dependency relations are generally useful~\citep{felice-yuan-2014-err,leacock2014automated,rozovskaya-roth-2014-building,wang2020comprehensive}. The details of training vary depending upon the classification algorithm, but popular examples include naive Bayes~\citep{rozovskaya-roth-2011-algorithm,kochmar-etal-2012-hoo}, maximum entropy~\citep{lee-2004-automatic,han_chodorow_leacock_2006,chodorow-etal-2007-detection,de-felice-2008-automatic}, decision trees~\citep{gamon-etal-2008-using}, support-vector machines~\cite{putra-szabo-2013-uds}, and the averaged perceptron~\citep{rozovskaya-roth-2010-generating,rozovskaya-roth-2010-training,rozovskaya-roth-2011-algorithm}.

More recently, neural network techniques have been applied to classification-based GEC, where neural classifiers have been built using context words with pre-trained word embeddings, like Word2Vec~\citep{mikolov-etal-2013-distributed} and GloVe~\citep{pennington-etal-2014-glove}. Different neural network models have been proposed, including convolutional neural networks (CNN)~\citep{sun-etal-2015-convolutional}, recurrent neural networks (RNN)~\citep{wang-etal-2017_deep,li-etal-2019-laix}, and pointer networks~\citep{li-etal-2019-laix}.

One limitation of these classifiers, however, is that they only target very specific error types with small confusion sets and do not extend well to errors involving open-class words (such as word choice errors). Another weakness is that they heavily rely on local context and treat errors independently, assuming that there is only one error in the context and all the surrounding information is correct. When multiple classifiers are combined for multiple error types, classifier order also matters and predictions from individual classifiers may become inconsistent~\citep{yuan-2017-grammatical}. These limitations consequently mean classifiers are generally no longer explored in GEC in favour of other methods. 

\subsection{Statistical Machine Translation}

\begin{figure}[t!]
	\begin{center}
		\scalebox{0.8}{
			\begin{tikzpicture}
			
			\draw (0,0) node[minimum height=1.6cm, minimum width=3.2cm, draw, thick, align=center](source) {\textbf{Source}\\$\mbox{P(C)}$}; 
			
			\draw (5,0) node[minimum height=1.6cm, minimum width=3.2cm, draw, thick, align=center](channel) {\textbf{Noisy channel}\\$\mbox{P(E}|\mbox{C)}$}; 
			
			\draw (10,0) node[minimum height=1.6cm, minimum width=3.2cm, draw, thick, align=center](receiver) {\textbf{Receiver}}; 
			
			\draw [->,thick,-stealth] (source) -- (channel);
			\draw (source) edge node[above] {$\mbox{C}$} (channel);
			
			\draw [->,thick,-stealth] (channel) -- (receiver);
			\draw (channel) edge node[above] {$\mbox{E}$} (receiver);
			
			\draw [->,thick,-stealth] (receiver) -- (12.8,0);
			\draw (receiver) edge node[above] {$\hat{\mbox{C}}$} (12.8,0);
			
			\end{tikzpicture}}
			\caption{\label{fig:noisy}The noisy channel model~\citep{shannon-1948-mathematical}.}
	\end{center}
\end{figure}

In contrast with statistical classifiers, one of the main advantages treating GEC as a statistical machine translation (SMT) problem is that SMT can theoretically correct all error types simultaneously without expert knowledge or feature engineering. This also includes interacting errors, which are problematic for rule-based systems and classifiers. Despite originally being developed for translation between different languages, SMT has been successfully applied to GEC, which can be seen as a translation problem from errorful to correct sentences. More specifically, although both the source and target sentences are in the same language, i.e. monolingual translation, the source may contain grammatical errors which should be `translated' to appropriate corrections. SMT is inspired by the \emph{noisy channel model}~\citep{shannon-1948-mathematical} and is mathematically formulated using Bayes' rule:
\begin{equation}
\label{equation:noisy}
\hat{C}=\argmax_{C}P(C|E)= \argmax_{C}\frac{P(E|C)P(C)}{P(E)}= \argmax_{C}P(E|C)P(C)
\end{equation}

\noindent where a correct sentence $C$ is said to have passed through a noisy channel to produce an erroneous sentence $E$, and the goal is to reconstruct the correct sentence $\hat{C}$ using a language model (LM) $P(C)$ and a translation model (TM) $P(E|C)$ - see \autoref{fig:noisy}. Candidate sentences are generated by means of a decoder, which normally uses a beam search strategy. The denominator $P(E)$ in~\autoref{equation:noisy} is ignored since it is constant across all $C$s.

The use of SMT for GEC was pioneered by~\citet{brockett-etal-2006-correcting}, who built a system to correct errors involving 14 countable and uncountable nouns. Their training data comprised a large corpus of sentences extracted from news articles which were deliberately modified to include artificial mass noun errors. \citet{mizumoto-etal-2011-mining} applied the same techniques to Japanese error correction but improved on them by not only considering a wider set of error types, but also training on real learner examples extracted from the language learning social network website Lang-8. \citet{yuan-felice-2013-constrained} subsequently trained a POS-factored SMT system to correct five types of errors in learner text for the CoNLL-2013 shared task, and revealed the potential of using SMT as a general approach for correcting multiple error types and interacting errors simultaneously. In the following year, the two top-performing systems in the CoNLL-2014 shared task demonstrated that SMT yielded state-of-the-art performance on general error correction in contrast with other methods~\citep{felice-etal-2014-grammatical,junczys-dowmunt-grundkiewicz-2014-amu}. This success led to SMT becoming a dominant approach in the field and inspired other researchers to adapt SMT technology for GEC, including:

\begin{itemize}

    \item Adding GEC-specific features to the model to allow for the fact that most words translate into themselves and errors are often similar to their correct forms. Two types of these features include the Levenshtein distance~\citep{felice-etal-2014-grammatical,junczys-dowmunt-grundkiewicz-2014-amu,junczys-dowmunt-grundkiewicz-2016-phrase,yuan-etal-2016-candidate,grundkiewicz-junczys-dowmunt-2018-near} and edit operations~\citep{junczys-dowmunt-grundkiewicz-2016-phrase,chollampatt-ng-2017-connecting,grundkiewicz-junczys-dowmunt-2018-near}.
    
    \item Tuning parameter weights with different algorithms, including minimum error rate training (MERT)~\citep{kunchukuttan-etal-2014-tuning,junczys-dowmunt-grundkiewicz-2014-amu}, the margin infused relaxed algorithm (MIRA)~\citep{junczys-dowmunt-grundkiewicz-2014-amu}, and pairwise ranking optimization (PRO)~\citep{junczys-dowmunt-grundkiewicz-2016-phrase}.
    
    \item Training additional large-scale LMs on monolingual native data, such as the British National Corpus (BNC)~\citep{yuan-2017-grammatical}, Wikipedia~\citep{junczys-dowmunt-grundkiewicz-2014-amu,chollampatt-ng-2017-connecting}, and Common Crawl~\citep{junczys-dowmunt-grundkiewicz-2014-amu,junczys-dowmunt-grundkiewicz-2016-phrase,chollampatt-ng-2017-connecting}.
    
    \item Introducing neural network components, such as a neural network global lexicon model (NNGLM) and neural network joint model (NNJM) ~\citep{chollampatt-etal-2016-neural, chollampatt-ng-2017-connecting}.
    
\end{itemize}

Despite their success in GEC, SMT-based approaches suffer from a few shortcomings. In particular, they i) tend to produce locally well-formed phrases with poor overall grammar, ii) exhibit a predilection for changing phrases to more frequent versions even when the original is correct, resulting in unnecessary corrections, iii) are unable to process long-range dependencies and iv) are hard to constrain to particular error types~\citep{felice-2016-artificial,yuan-2017-grammatical}. Last but not least, the performance of SMT systems depends heavily on the amount and quality of parallel data available for training, which is very limited in GEC. A common solution to this problem is to generate artificial datasets, where errors are injected into well-formed text to produce pseudo-incorrect sentences, as described in~\autoref{sec:aeg}.

\subsection{Neural Machine Translation}

With the advent of deep learning and promising results reported in machine translation and other sequence-to-sequence tasks, neural machine translation (NMT) was naturally extended to GEC. Compared to SMT, NMT uses a single large neural network to model the entire correction process, eliminating the need for complex GEC-specific feature engineering. Training an NMT system is furthermore an end-to-end process and so does not require separately trained and tuned components as in SMT. Despite its simplicity, NMT has achieved state-of-the-art performance on various GEC tasks \citep{flachs-etal-2021-data, rothe-etal-2021-simple}. 

NMT employs the \emph{encoder--decoder} framework~\citep{cho-etal-2014-learning}. An encoder first reads and encodes an entire input sequence $\mathbf{x}=(x_1, x_2, ..., x_T)$ into hidden state representations, and a decoder then generates an output sequence $\mathbf{y}=(y_1, y_2, ..., y_{T'})$ by predicting the next word $y_t$ based on the input sequence $\mathbf{x}$ and all the previously generated words $\{y_1, y_2, ..., y_{t-1}\}$:
\begin{equation}
p(\mathbf{y}|\mathbf{x})=\prod_{t=1}^{T'}p(y_t|\{y_1, y_2, ..., y_{t-1}\}, \mathbf{x})
\end{equation}

Different network architectures have been proposed for building the encoders and decoders; three commonly used sequence-to-sequence models are RNNs~\cite{bahdanau-etal-2015-neural}, CNNs~\citep{gehring-etal-2017-convolutional}, and Transformers~\citep{transformer}.

\subsubsection{Recurrent Neural Networks} \hfill \\

\noindent Recurrent Neural Networks (RNN) are a type of neural network that is specifically designed to process sequential data. RNNs are used to transform a variable-length input sequence to another variable-length output sequence~\citep{cho-etal-2014-learning,sutskever-etal-2014-sequence}. To handle long-term dependencies, gated units are usually used in RNNs \citep{Goodfellow-et-al-2016}. The two most effective RNN gates are Long-Short Term Memory (LSTM)~\citep{hochreiter-schmidhuber-1997-long} and Gated Recurrent Units (GRU)~\citep{cho-etal-2014-learning}. \citet{bahdanau-etal-2015-neural} introduced an attention mechanism to implement variable-length representations, which eased optimisation difficulty and resulted in improved performance. \citet{yuan-briscoe-2016-grammatical} presented the first work on NMT-based approach for GEC. Their model consists of a bidirectional RNN encoder and an attention-based RNN decoder. \citet{xie-etal-2016-neural} proposed the use of a character-level RNN sequence-to-sequence model for GEC. Following their work, a hybrid model with nested attention at both the word and character level was later introduced by~\citet{ji-etal-2017-nested}.

\subsubsection{Convolutional Neural Networks} \hfill \\

\noindent Another way of processing sequential data is by using a convolutional neural network (CNN) across a temporal sequence. CNNs are a type of neural network that is designed to process grid-like data and specialises in capturing local dependencies \citep{Goodfellow-et-al-2016}. CNNs were first applied to NMT by \citet{kalchbrenner-blunsom-2013-recurrent}, but they were not as successful as RNNs until~\citet{gehring-etal-2017-convolutional} stacked several CNN layers followed by non-linearities. Inspired by this work, \citet{chollampatt-ng-2018-multilayer} proposed a 7-layer CNN sequence-to-sequence model for GEC. In their model, local context is captured by the convolution operations performed over smaller windows and wider context is captured by the multi-layer structure. Their model was the first NMT-based model that significantly outperformed prior SMT-based models. This model was later used in combination with Transformers to build a state-of-the-art GEC system~\citep{yuan-etal-2019-neural}.

\subsubsection{Transformers}\label{sec:transformer} \hfill \\

\noindent The Transformer \cite{transformer} is the first sequence transducer network that entirely relies on a self-attention mechanism to compute the representations of its input, without the need for recurrence or convolution. Its architecture allows better parallelisation on multiple GPUs, overcoming the weakness of RNNs.

The Transformer has become the architecture of choice for machine translation since its inception \cite{edunov-etal-2018-understanding,wang2018multiagent,deep_tf_mt}. Previous work has investigated the adaptation of NMT to GEC, such as optimising the model with edit-weighted loss \citep{junczys-dowmunt-etal-2018-approaching} and adding a copy mechanism ~\citep{zhao-etal-2019-improving,yuan-etal-2019-neural}. A copy mechanism allows the model to directly copy tokens from the source sentence, which often has substantial overlap with the target sentence in GEC. The Copy-Augmented Transformer has become a popular alternative architecture for GEC \citep{hotate-etal-2020-generating, wan-etal-2020-improving}. Another modification to the Transformer architecture is altering the encoder-decoder attention mechanism in the decoder to accept and make use of additional context. For example, \citet{kaneko-etal-2020-encoder} added the BERT representation of the input sentence as additional context for GEC, while \citet{yuan-bryant-2021-document} added the previous sentences in the document, and \citet{zhang-etal-2022-syngec} added a tree-based syntactic representation of the input sentence.

As the Transformer architecture has a large number of parameters, yet parallel GEC training data is limited, pre-training has become a standard procedure in building GEC systems. The first Transformer-based GEC system \citep{junczys-dowmunt-etal-2018-approaching} pre-trained the Transformer decoder on a language modeling task, but it has since become more common to pre-train on synthetic GEC data. The top two systems in the BEA-2019 shared task \citep{grundkiewicz-etal-2019-neural, choe-etal-2019-neural} and a recent state-of-the-art GEC system \citep{stahlberg-kumar-2021-synthetic} both pre-trained their Transformer models with synthetic data, but they generated their synthetic data in different ways. We discuss different techniques for generating synthetic data in \autoref{subsec:synthetic_data}. More recently, with the advances in large pre-trained language models, directly fine-tuning large pre-trained language models with GEC parallel data has been shown to achieve comparable performance with synthetic data pre-training \citep{katsumata-komachi-2020-stronger}, even reaching state-of-the-art performance \citep{rothe-etal-2021-simple, tarnavskyi-etal-2022-ensembling}.

Irrespective of the type of NMT architecture however (RNN, CNN, Transformer), NMT systems share several weaknesses with SMT systems, most notably in terms of data requirements. In particular, although NMT systems are more capable at correcting longer range and more complex errors than SMT, they also require as much training data as possible, which can lead to extreme resource and time requirements: it is not uncommon for some models to require several days of training time on a cluster of GPUs. Moreover, neural models are almost completely uninterpretable (which furthermore makes them difficult to customise) and it is nearly impossible for a human to determine the reasoning behind a given decision; this is particularly problematic if we also want to explain the cause of an error to a user rather than just correct it. Ultimately however, a key strength of NMT is that it is an end-to-end approach, and so does not require feature engineering or much human intervention, and it is undeniable that it produces some of the most convincing output to date.  

\subsection{Edit-based approaches}\label{sec:seq2edits}

While most GEC approaches generate a corrected sentence from an input sentence, the edit generation approach generates a sequence of edits to be applied to the input sentence instead. As GEC has a high degree of token copying from the input to the output, \citet{stahlberg-kumar-2020-seq2edits} argued that generating the full sequence is wasteful. By generating edit operations instead of all tokens in a sentence, the edit generation approach typically has a faster inference speed, reported to be five to ten times faster than GEC systems that generate the whole sentence. One limitation of this approach, however, is that edit operations tend to be token-based, and so sometimes fail to capture more complex, multi-token fluency edits \citep{lai-etal-2022-type}. Edit generation has been cast as a sequence tagging task \cite{malmi-etal-2019-encode,awasthi-etal-2019-parallel,omelianchuk-etal-2020-gector,tarnavskyi-etal-2022-ensembling} or a sequence-to-sequence task \cite{stahlberg-kumar-2020-seq2edits}.

In the sequence tagging approach, for each token of an input sentence, the system predicts an edit operation to be applied to that token (\autoref{tab:edit_gen_seqlab}). This approach requires the user to define a set of tags representing the edit operations to be modelled by the system. Some edits can be universally modelled, such as conversion of verb forms or conversion of nouns from singular to plural form. Some others such as word insertion and word replacement are token-dependent. Token-dependent edits need a different tag for each possible word in the vocabulary, resulting in the number of tags growing linearly with the number of unique words in the training data. Thus, the number of token-dependent tags to be modelled in the system becomes a trade-off between coverage and model size.

\begin{table}[h]
\centering
\begin{adjustbox}{width=\textwidth}
\setlength{\tabcolsep}{3pt}
 \begin{tabular}{l|llllllllllll} 
 \hline
Source & After & many & years & {} & he & still & dream & to & become & a & super & hero\\
Target & After & many & years &, & he & still & dreams & of & becoming & a & super & hero\\ 
\small Edits & \small KEEP & \small KEEP & \small APP\_, & {} & \small KEEP & \small KEEP & \small VB\_VBZ & \small REP\_of & \small VB\_VBG & \small KEEP & \small KEEP & \small KEEP \\
\end{tabular}
\end{adjustbox}
\caption{Example task formulation of edit generation in the sequence tagging approach from \cite{omelianchuk-etal-2020-gector}. APP\_x denotes an operation of appending token x, and REP\_x denotes replacing the current token with x.}\label{tab:edit_gen_seqlab}
\end{table}

On the other hand, the sequence-to-sequence approach is more flexible as it does not limit the output to pre-defined edit operation tags. It produces a sequence of edits, each consisting of a span position, a replacement string, and an optional tag for edit type (\autoref{tab:edit_gen_seq2seq}). These tags add interpretability to the process and have been shown to improve model performance. As generation in the sequence-to-sequence approach has a left-to-right dependency, the inference procedure is slower than that in the sequence tagging approach. It is still five times faster than that in the whole sentence generation approach as the edit sequence generated is much shorter than the sequence of all tokens in the sentence \cite{stahlberg-kumar-2020-seq2edits}.

\begin{table}[h]
\centering
 \begin{tabular}{l| p{0.85\linewidth} } 
 \hline
Source & After many years he still dream to become a super hero .\\
Target & After many years , he still dreams of becoming a super hero .\\ 
Edits & (SELF,3,SELF), (PUNCT,3,`,'), (SELF,5,SELF), (SVA,6,`dreams'), (PART,7,`of'), (FORM,8,`becoming'), (SELF,12,SELF) \\
\end{tabular}
\caption{Example task formulation of edit generation in the sequence-to-sequence approach from \cite{stahlberg-kumar-2020-seq2edits}. Each tuple represents a tag, a span's ending position, and a replacement string.}
\label{tab:edit_gen_seq2seq}
\end{table}

The main advantages of edit-based approaches to GEC are thus that they not only add much needed transparency and explainability to the correction process, but they are also much faster at inference time than NMT. Their main disadvantages, however, are that they generally require human engineering to define the size and scope of the edit label set, and that it is more difficult to represent interacting and complex multi-token edits with token-based labels. Like all neural approaches, they also depend on as much training data as possible, but when data is available, edit-based approaches are very competitive with state-of-the-art NMT models.

\subsection{Language Models for Low-Resource and Unsupervised GEC}\label{sec:lm}

Unlike previous strategies, language model based GEC does not require training a system with parallel data. Instead, it employs various techniques using n-gram or Transformer language models. LM-based GEC was a common approach before machine translation-based GEC became popular \cite{dahlmeier-ng-2012-beam,lee-lee-2014-postech}, but has experienced a recent resurgence with low-resource GEC and unsupervised GEC due to the effectiveness of large Transformer-based language models \citep{alikaniotis-raheja-2019-unreasonable,grundkiewicz-junczys-dowmunt-2019-minimally,flachs-etal-2019-noisy}. Recent advances have enabled Transformer-based language models to more adequately capture syntactic phenomena \cite{jawahar-etal-2019-bert,wei-etal-2021-frequency}, making them capable GEC systems when little or no data is available. These systems can, however, become even more capable when exposed to a small amount of parallel data \citep{mita-yanaka-2021-grammatical}.

\subsubsection{Language models as Discriminators} \hfill \\

The traditional LM-based approach to GEC makes the assumption that low probability sentences are more likely to contain grammatical errors than high probability sentences, and so a GEC system must determine how to transform the former into the latter based on language model probabilities \cite{bryant-briscoe-2018-language}. Correction candidates can be generated from confusion sets \cite{dahlmeier-ng-2011-correcting,bryant-briscoe-2018-language}, classification-based GEC models \cite{dahlmeier-ng-2012-beam}, or finite-state transducers \cite{stahlberg-etal-2019-neural}.

\citet{yasunaga-etal-2021-lm} proposed an alternative method using the break-it-fix-it (BIFI) approach \citep{yasunaga2021break}, with a language model as the critic (LM-critic). Specifically, BIFI trains a breaker (noising channel) and a fixer (GEC model) on multiple rounds of feedback loops. An initial fixer is used to correct erroneous text, then the sentence pairs are filtered using LM-critic. Using this filtered data, the breaker is trained and used to generate new synthetic data from a clean corpus. These new sentence pairs are then also filtered using LM-critic and subsequently used to train the fixer again. The BIFI approach can be used for unsupervised GEC by training the fixer on synthetic data.

\subsubsection{Language models as Generators} \hfill \\

A more recent LM-based approach to GEC is to use a language model as a zero-shot or few-shot generator to generate a correction given a prompt and noisy input sentence. For example, given the prompt ``Correct the grammatical errors in the following text:'' followed by an input sentence, the language model is expected to generate a corrected form of the input sentence given the prompt as context. This approach has become possible largely due to the advent of Large Language Models (LLMs), such as GPT-2 \citep{radford2019language}, GPT-3 \citep{brown-2020-gpt3}, OPT \citep{zhang2022opt} and PaLM \citep{chowdhery2022palm}, which have been trained on up to a trillion words and parameterised using tens or hundreds of billions of parameters. These models have furthermore been shown to be capable of generalising to new unseen tasks or languages by being fine-tuned on a wide variety of other NLP tasks \citep{sanh-2022-multitask, wei-2022-finetuned, muennighoff2022crosslingual}, and so it is possible, for the first time, to build a system that is capable of carrying out multilingual GEC without having been explicitly trained to do so. 

Despite their potential however, there have not yet been any published studies that have formally benchmarked generative LLMs against any of the standard GEC test sets. Although a number of studies were beginning to appear at the time of final submission of this survey paper, most only evaluated LLM performance on a small sample (100 sentences) of the official test sets \citep{wu2023chatgpt,coyne2023analysis}. These studies generally conclude, however, that LLMs have a tendency to overcorrect for fluency, which causes them to underperform on datasets that were developed for minimal corrections \citep{fang2023chatgpt}. We expect further investigation of this phenomenon in the coming year.

Regardless of the type of language model, the main advantage of language model based approaches is that they only require unannotated monolingual data and so are much easier to extend to other languages than all other approaches. While discriminative LMs may not perform as well as state-of-the-art models and generative LLMs models have not been formally benchmarked, LMs have nevertheless proven themselves capable and can theoretically correct all types of errors, including complex fluency errors. The main disadvantage of language model approaches, however, is that it can be hard to adequately constrain the model, and so models sometimes replace grammatical words with other words that simply occur more frequently in a given context. An additional challenge in generative LLM-based GEC is that prompt engineering is important \citep{liu-2023-prompt} and output may vary depending on whether a system was asked to `correct' a grammatical error or `fix' a grammatical error \citep{coyne2023analysis}. Ultimately, all LM-based approaches suffer from the limitation that probability is not grammaticality, and so rare words may be mistaken for errors. 

\section{Additional Techniques}
\label{sec:add_tech}

While \autoref{sec:core_app} introduced the core technologies underpinning modern GEC systems, a number of other techniques are also commonly applied to further boost performance. Several of these techniques are introduced in this section, including re-ranking, ensembling and system combination, multi-task learning, custom inference methods (e.g., iterative decoding), contextual GEC, and Generative Adversarial Networks (GANs).  

\subsection{Re-ranking}
\label{sec:reranking}

Machine translation based (both SMT and NMT) systems can produce an $n$-best list of alternative corrections for a single sentence. This has led to much work on $n$-best list re-ranking, which aims to determine whether the best correction for a sentence is not the most likely candidate produced by the system (i.e. $n=1$), but is rather somewhere further down the top $n$ most likely candidates~\citep{yuan-etal-2016-candidate,mizumoto-matsumoto-2016-discriminative,hoangijcai16}. As a separate post-processing step, candidates produced by an SMT-based or NMT-based GEC system can be re-ranked using a rich set of features that have not been explored by the decoder before, so that better candidates can be selected as `optimal' corrections. During re-ranking, GEC-specific features can then be easily adapted without worrying about fine-grained model smoothing issues. In addition to the original model scores of the candidates, useful features include:

\begin{itemize}
    
    \item sentence fluency scores calculated from: LMs~\citep{yuan-etal-2016-candidate,chollampatt-ng-2018-multilayer}, neural error detection models~\citep{yannakoudakis-etal-2017-neural,yuan-etal-2019-neural}, neural quality estimation models ~\citep{chollampatt-ng-2018-neural}, and BERT~\citep{kaneko-etal-2019-tmu};
    
    \item similarity measures like Levenshtein Distance~\citep{yannakoudakis-etal-2017-neural,yuan-etal-2019-neural} and edit operations~\citep{chollampatt-ng-2018-multilayer,kaneko-etal-2019-tmu};
    
    \item length-based features~\citep{yuan-etal-2016-candidate};
    
    \item right-to-left models~\citep{grundkiewicz-etal-2019-neural,kaneko-etal-2020-encoder};
    
    \item syntactic features like POS n-grams, dependency relations~\citep{mizumoto-matsumoto-2016-discriminative};
    
    \item error detection information which has been used in a binary setting~\citep{yannakoudakis-etal-2017-neural,yuan-etal-2019-neural}, as well as a multi-class setting~\citep{yuan-etal-2021-multi}.
    
\end{itemize}

$N$-best list reranking has traditionally been one of the simplest and most popular methods of boosting system performance. An alternative form of reranking is to collect all the edits from the $N$-best corrections and filter them using an edit-scorer \cite{sorokin-2022-improved}.

\subsection{Ensembling and System Combination}
\label{sec:sys_comb}

Ensembling is a common technique in machine learning to combine the predictions of multiple individually trained models. Ensembles often generate better predictions than any of the single models that are combined \cite{opitz-maclin-1999-popular}. In GEC, ensembling usually refers to averaging the probabilities of individually trained GEC models when predicting the next token in the sequence-to-sequence approach or the edit tag in the edit-based approach. GEC models that are combined into ensembles usually have similar properties with only slight variations, which can be the random seed \cite{stahlberg-kumar-2021-synthetic}, the pre-trained model \cite{omelianchuk-etal-2020-gector}, or the architecture \cite{choe-etal-2019-neural}.

On the other hand, different GEC approaches have different strengths and weaknesses. \citet{susanto-etal-2014-system} have shown that combining different GEC systems can produce a better system with higher accuracy. When combining systems that have substantial differences, training a system combination model is preferred over ensembles. A system combination model allows the combined system to properly integrate the strengths of the GEC systems and has been shown to produce better results than ensembles \cite{kantor-etal-2019-learning,qorib22}. The combination model can be trained through learning the characteristic of the GEC systems \cite{kantor-etal-2019-learning,gec-ip,qorib22} or learning how to score a correction by supplying the model with examples of good and bad corrections for different kinds of student sentences \cite{sorokin-2022-improved}. Moreover, most system combination methods for GEC work on a \textit{black-box} setup \cite{kantor-etal-2019-learning,gec-ip,qorib22}, only requiring the systems' outputs without any access to the systems' internals and the prediction probabilities. When the individual component systems are not different enough, encouraging the individual systems to be more diverse before combining them can also improve performance \cite{ddc}.

\subsection{Multi-task learning}
\label{sec:multi-learn}

Multi-task learning allows systems to use information from related tasks and learn from multiple objectives via shared representations, leading to performance gains on individual tasks. \citet{rei-yannakoudakis-2017-auxiliary} was the first to investigate the use of different auxiliary objectives for the task of error detection in learner writing through a neural sequence-labelling model. In addition to predicting the binary error labels (i.e. correct or incorrect), they experimented with also predicting specific error type information, including the learner's L1, token frequency, POS tags and dependency relations. \citet{asano-etal-2019-aip} employed a similar approach in which their error correction model additionally estimated the learner's language proficiency level and performed sentence-level error detection simultaneously. Token-level and sentence-level error detection have also both been explored as auxiliary objectives in NMT-based GEC~\citep{yuan-etal-2019-neural,zhao-etal-2019-improving}, where systems have been trained to jointly generate a correction and predict whether the source sentence (or any token in it) is correct or incorrect. Labels for these auxiliary error detection tasks can be extracted automatically from existing datasets using automatic alignment tools like ERRANT~\citep{bryant-etal-2017-automatic}.

\subsection{Custom inference methods}

Various inference techniques have been proposed to improve the quality of system output or speed up inference time in GEC. The most common of these, which specifically improves output quality, is to apply multiple rounds of inference, known as \textit{iterative decoding} or \textit{multi-turn decoding}. Since the input and output of GEC are in the same language, the output of the model can be passed through the model again to produce a second iteration of output. The advantage of this is that the model gets a second chance to correct errors it might have missed during the first iteration. \citet{lichtarge-etal-2019-corpora} thus proposed an iterative decoding algorithm that allows a model to make multiple incremental corrections. In each iteration, the model is allowed to generate a different output only if it has high confidence. This technique proved effective for GEC systems trained on noisy data such as Wikipedia edits, but not as effective on GEC systems trained on clean data. \citet{ge-etal-2018-fluency} proposed an alternative iterative decoding technique called fluency boost, in which the model performs multiple rounds of inference until a fluency score stops increasing, while \citet{lai-etal-2022-type} proposed an iterative approach that investigated the effect of correcting different types of errors (missing, replacement, unnecessary words) in different orders. Iterative decoding is commonly employed in sequence-labelling GEC systems which cannot typically correct all errors in a single pass. In these systems, iterative decoding is applied until the model stops making changes to the output or the number of iterations reaches a limit \citep{awasthi-etal-2019-parallel,omelianchuk-etal-2020-gector,tarnavskyi-etal-2022-ensembling}.

Other inference techniques have been proposed to speed up inference time in GEC. As many tokens in GEC are copied from the input to the output, standard left-to-right inference can be inefficient. \citet{chen-etal-2020-improving-efficiency} thus proposed a two-step process that only performs correction on text spans that are predicted to contain grammatical errors. Specifically, their system first predicts erroneous spans using an erroneous span detection (ESD) model, and then corrects only the detected spans using an erroneous span correction (ESC) model. They reported reductions in inference time of almost 50\% compared to a standard sequence-to-sequence model. In contrast, \citet{sun-etal-2021-instantaneous} proposed a parallelisation technique to speed up inference, aggressive decoding, which can be applied to any sequence-to-sequence model. Specifically, aggressive decoding first decodes as many tokens as possible in parallel and then only re-decodes tokens one-by-one at the point where the input and predictions differ (if any). When the input and predicted tokens start to match again, aggressive decoding again decodes the remainder in parallel until either the tokens no longer match or the end-of-sentence token is predicted. Since the input and output sequences in GEC are often very similar, this means most tokens can be decoded aggressively, yielding an almost ten time speedup in inference time.


\subsection{Contextual GEC}

Context provides valuable information that is crucial for correcting many types of grammatical errors and resolving inconsistencies. Existing GEC systems typically perform correction at the sentence-level however, i.e. each sentence is processed independently, and so cross-sentence information is ignored. These systems thus frequently fail to correct contextual errors, such as verb tense, pronoun, run-on sentence and discourse errors, which typically rely on information outside the scope of a single sentence. Corrections proposed by such narrow systems are furthermore likely to be inconsistent throughout a paragraph or entire document. 

\citet{chollampatt-etal-2019-cross} were the first to address this problem by adapting a CNN sequence-to-sequence model to be more context-aware. Specifically, they introduced an auxiliary encoder to encode the two previous sentences along with the input sentence and incorporated the encoding in the decoder via attention and gating mechanisms. \citet{yuan-bryant-2021-document} subsequently compared different architectures for capturing wider context in Transformer-based GEC and showed that local context is useful ($\leq2$ sentences) but very long context ($>2$ sentences) is not necessary for improved performance.

Since human reference edits are not annotated for whether an error depends on local context or long range context, it is often difficult to evaluate the extent to which a context-aware system improves the correction of context-sensitive errors. \citet{chollampatt-etal-2019-cross} thus constructed a synthetic dataset of verb tense errors which required cross-sentence context for correction, and \citet{yuan-bryant-2021-document} proposed a document-level evaluation framework to address this problem.

\subsection{Generative Adversarial Networks}\label{sec:gan}

Generative Adversarial Networks (GANs) \citep{NIPS2014_5ca3e9b1} are an approach to model training that makes use of both a generator, to generate some output; and a discriminator, to discriminate between real data and artificial output. In the context of GEC, \citet{raheja-alikaniotis-2020-adversarial} were the first to apply this methodology to error correction, in which they trained a standard sequence-to-sequence Transformer model to generate grammatical sentences from parallel data (the generator) and a sentence classification model to discriminate between these generated output sentences and human-annotated reference sentences (the discriminator). During training, the models competed adversarially such that the generator learnt to generate corrected sentences that are indistinguishable from the reference sentences (and thus fooled the discriminator), while the discriminator learnt to identify the differences between real and generated sentences (and thus defeated the generator). This adversarial training process was ultimately shown to produce a better sequence-to-sequence model.

In addition to sequence-to-sequence generation, GANs have also been applied to sequence-labelling for GEC. In particular, \citet{parnow-etal-2021-grammatical} trained a generator to generate increasingly realistic errors (in the form of token-based edit labels) and a discriminator to differentiate between artificially-generated edits and real human edits. They similarly reported improvements over a baseline that was not trained adversarially. 



\section{Data Augmentation}
\label{sec:aeg}

A common problem in GEC is that the largest publicly-available high-quality parallel corpora only contain roughly 50k sentence pairs, and larger corpora, such as Lang-8, are noisy \citep{mita-etal-2020-self,rothe-etal-2021-simple}. This data sparsity problem has motivated a lot of research into synthetic data generation, especially in the context of resource-heavy NMT approaches, because synthetic data primarily requires a native monolingual source corpus rather than a labour-intensive manual annotation process. In this section, we introduce several different types of data augmentation methods, including rule-based noise injection and back-translation, but also noise reduction which aims to improve the quality of existing datasets by removing/down-weighting noisy examples. It is an open question as to how to best evaluate the quality of synthetic data \citep{htut-tetreault-2019-unbearable,white-rozovskaya-2020-comparative}. An effort has been made by \cite{kiyono-etal-2019-empirical} to compare the noise injection method and back-translation, but it is hard to comprehensively compare synthetic data generation methods directly, so most research evaluates it indirectly in terms of its impact on the performance of previous experiments. Data augmentation has nevertheless contributed greatly to GEC system improvement and has become a staple component of recent models.

\subsection{Synthetic Data Generation}\label{subsec:synthetic_data}

GEC is sometimes regarded as a low-resource machine translation task \citep{junczys-dowmunt-etal-2018-approaching}. With the dominance of neural network approaches, the need for more data grows as model size continues to increase. However, obtaining human annotations is expensive and difficult. Thus, techniques to generate synthetic parallel corpora from clean monolingual corpora have been intensely explored. A synthetic parallel corpus is generated by adding noise to a sentence and pairing it with the original sentence. The corrupted sentence is then regarded as a learner’s sentence (source) and the original clean sentence is regarded as the reference (target). There are many ways to generate synthetic sentences, and the dominant techniques usually fall under the category of noise injection or back-translation \cite{kiyono-etal-2019-empirical}.

\subsubsection{Noise Injection}\label{sec:noise-injection} \hfill \\

\noindent One way to artificially generate grammatical errors to clean monolingual corpora is by perturbing a clean text to make it grammatically incorrect. The perturbations can be in the form of rule-based noising operations or error patterns that usually appear in GEC parallel corpora.

\paragraph{Rule-based} The most intuitive way of adding noise to a clean corpus is by applying a series of perturbation operations based on some pre-defined rules. The rules are applied based on a probability, which can be decided arbitrarily, empirically, or through some observations of available data. \citet{ehsan2013grammatical} apply one error to each sentence from pre-defined error templates that include omitting prepositions, repeating words, and so on. \citet{lichtarge-etal-2019-corpora} introduce spelling errors to Wikipedia edit history by performing deletion, insertion, replacement, and transposition of characters. \citet{zhao-etal-2019-improving} also apply a similar noising strategy but at the word level, that is deleting, adding, shuffling, and replacing words in a sentence. \citet{grundkiewicz-etal-2019-neural} combine both approaches, character-level and word-level noising, but word substitution is limited to pairs from a confusion set made from an inverted spellchecker. Similarly, \citet{xu-etal-2019-erroneous} also combine both approaches but with a more complex word substitution strategy by making use of part-of-speech (POS) tags. The rule-based injection technique can also be applied dynamically during training to increase the error rate in a parallel corpus instead of creating additional training data \citep{Zhao_Wang_2020}.

\paragraph{Error patterns} Another way of generating synthetic data is through injecting errors that frequently occur in GEC parallel corpora. In this way, the errors are more similar to the ones that humans usually make. \citet{rozovskaya-roth-2010-training} proposed three different methods of injecting article errors, based on the error distribution in English as a Second Language (ESL) data. They proposed adding article errors based on the distribution of articles in a text before correction, the distribution of articles in the corrected text, and the distribution of article corrections themselves. \citet{felice-yuan-2014-generating} later improved the method by taking into consideration the morphology, POS tag, semantic concept, and word sense information of a text when generating the artificial errors. \citet{rei-etal-2017-artificial} further extended it to all types of errors. Another direction of emulating human errors is by extracting the correction patterns from GEC parallel corpora and applying the inverse of those corrections on grammatically correct sentences, as done by \citet{yuan-felice-2013-constrained} using the corrections from the NUCLE corpus and by \citet{choe-etal-2019-neural} using the corrections from the W\&I training data. The correction patterns are extracted both in lexical form (an → the) and POS (NN → NNS).

\pagebreak
\subsubsection{Back-translation} 
\label{sec:back-translation}
\hfill \\

\noindent Emulating human errors can be made in a more automated and dynamic way via a noisy channel model. The noisy channel model is trained with the inverse of a GEC parallel corpus, treating the learner’s sentence as the target and the reference sentence as the source. This technique is commonly called back-translation. The technique was originally proposed for generating additional data in machine translation \citep{sennrich-etal-2016-improving}, but it is also directly applicable to GEC. \citet{rei-etal-2017-artificial} were the first to apply back-translation to grammatical error detection (GED) and \citet{xie-etal-2018-noising} were the first to apply it to GEC. \citet{yuan-etal-2019-neural} add a form of quality control to \citet{rei-etal-2017-artificial} based on language model probabilities in an effort to make sure that the generated synthetic sentences are less probable (and hence hopefully less grammatical) than the original input sentences. Between the rule-based and back-translation strategy, \citet{kiyono-etal-2019-empirical} report that the back-translation strategy has better empirical performance. They also compare back-translation with a noisy beam-search strategy \citep{xie-etal-2018-noising} and back-translation with sampling strategy \citep{edunov-etal-2018-understanding}, and report that both achieve competitive performance. \citet{koyama-etal-2021-comparison} furthermore compare the effect of using different architectures (e.g. CNN, LSTM, Transformer) for back-translation, and find that interpolating multiple generation systems tends to produce better synthetic data to train a GEC system on. Another variant of back-translation was proposed by \citet{stahlberg-kumar-2021-synthetic} to generate more complex edits. They found that generating a sequence of edits using Seq2Edit \citep{stahlberg-kumar-2020-seq2edits} works better than generating the corrupted sentences directly. They also reported that back-translation with sampling worked better than beam search in their experiments.

\subsubsection{Round-trip Translation} \hfill \\

\noindent A less popular alternative to back-translation is round-trip translation, which generates synthetic sentence pairs via a bridge language; e.g. English-Chinese-English. The assumption is that the MT system will make translation errors and so the output via the bridge language will be noisy in relation to the input. This strategy was employed by \citet{Madnani-et-al-2012} and \citet{lichtarge-etal-2019-corpora}, who furthermore both explored the effect of using different bridge languages. \citet{zhou-etal-2020-improving-grammatical} explore a similar technique, except use a bridge language as the input to both a low-quality and high-quality translation system (namely SMT vs. NMT), and treat the output from the former as an ungrammatical noisy sentence and the output from the latter as the reference.

\subsection{Augmenting Official Datasets}\label{sec:aug_off}

Besides generating synthetic data to address the data sparsity problem in GEC, other works focus on augmenting official datasets, via noise reduction or model enhancement.

Noise reduction aims to reduce the impact of wrong corrections in the official GEC datasets. One direction focuses on correcting noisy sentences. \citet{mita-etal-2020-self} and \citet{rothe-etal-2021-simple} achieve this by incorporating a well-trained GEC model to reduce wrong corrections. The other direction attempts to down-weight noisy sentences. \citet{lichtarge-etal-2020-data} introduce an offline re-weighting method to score each training sentence based on delta-log perplexity, $\Delta ppl$, which measures the model's log perplexity difference between checkpoints for a single sentence. Sentences with lower $\Delta ppl$ are preferred and assigned a higher weight during training. 

Model enhancement augments official datasets to address the model's weakness.  \citet{parnow-etal-2021-grammatical} aim to enhance performance by reducing the error density mismatch between training and inference. They use a generative adversarial network (GAN) \cite{NIPS2014_5ca3e9b1} to produce an ungrammatical sentence that could better represent the error density at inference time. \citet{lai-etal-2022-type} also address the mismatch between training and inference, but specific to multi-round inference. They propose additional training stages that make the model consider edit type interdependence when predicting the corrections. \citet{cao-etal-2021-grammatical-error} aim to enhance model performance in low-error density domains. The augmented sentences are generated by beam search to capture wrong corrections that the model tends to make. Supervised contrastive learning \cite{pmlr-v119-chen20j} is then applied to enhance model performance. \citet{cao-etal-2022-eb} use augmented sentences generated during beam search to address the exposure bias problem in seq2seq GEC models. A dynamic data reweighting method through reinforcement learning is used to select an optimal sampling strategy for different beam search candidates.

\section{Evaluation}
\label{sec:eval}

A core component of any NLP system is the ability to measure model performance. This section hence first introduces the most commonly-used evaluation metrics in GEC, namely the MaxMatch (M$^2$) scorer \citep{dahlmeier2012}, ERRANT \citep{bryant-etal-2017-automatic,felice-etal-2016-automatic} and GLEU \citep{napoles2015,gleu+}, as well as other reference-based and reference-less metrics that have been proposed. It next discusses the problem of metric reliability, particularly in relation to correlation with human judgements, and explains why it is difficult to draw any robust conclusions. The section concludes with a discussion of best practices in GEC evaluation, including defining standard experimental settings and highlighting their limitations. To date, almost all evaluation in GEC has been carried out at the sentence level. 

\subsection{MaxMatch}

One of the most prevalent evaluation methods used in current GEC research is the MaxMatch (M$^2$) scorer\footnote{\url{https://www.comp.nus.edu.sg/~nlp/conll14st.html}} \citep{dahlmeier2012} which calculates an F$_\beta$-score \citep{f1}. Specifically, the M$^2$ scorer is a reference-based metric which compares system hypothesis edits against human-annotated reference edits and counts a True Positive (TP) if a hypothesis edit matches a reference edit, a False Positive (FP) if a hypothesis edit does \textit{not} match any reference edit, and a False Negative (FN) if a reference edit does \textit{not} match any hypothesis edit. An example of each case is shown below.

\begin{center}
\begin{tabular}{l|lllllll}
\multicolumn{1}{l}{} &  & \textbf{TP} & & \textbf{FN} & \textbf{FP} &  & \\
Original & I     & \textcolor[rgb]{ 0,  .69,  .314}{likes} & to    & \textcolor[rgb]{ .439,  .188,  .627}{drive} & \textcolor[rgb]{ 1,  0,  0}{a} & bicycle & . \\
Hypothesis & I     & \textcolor[rgb]{ 0,  .69,  .314}{like} & to    & \textcolor[rgb]{ .439,  .188,  .627}{drive} & \textcolor[rgb]{ 1,  0,  0}{the} & bicycle & . \\
Reference & I     & \textcolor[rgb]{ 0,  .69,  .314}{like} & to    & \textcolor[rgb]{ .439,  .188,  .627}{ride} & \textcolor[rgb]{ 1,  0,  0}{a} & bicycle & . \\
\end{tabular}%
\end{center}

The total number of TPs, FPs and FNs for a dataset can then be used to calculate Precision (P) (\autoref{p}) and Recall (R) (\autoref{r}), which respectively denote the proportion of hypothesis edits that were correct and the proportion of reference edits that were found in the hypothesis edits, which in turn can be used to calculate the F$_\beta$-score (\autoref{fbeta}). In current GEC research, it is common practice to use $\beta=0.5$, first introduced in \citep{conll2014}, which weights precision twice as much as recall, because it is generally considered more important for a GEC system to be precise than to necessarily correct all errors. 

\noindent
\begin{minipage}{.275\linewidth}
\leqnos
\begin{equation}
  P = \frac{TP}{TP+FP}
  \label{p}
\end{equation}
\end{minipage}%
\begin{minipage}{.275\linewidth}
\leqnos
\begin{equation}
  R = \frac{TP}{TP+FN}
  \label{r}
\end{equation}
\end{minipage}
\begin{minipage}{.45\linewidth}
\leqnos
\begin{equation}
  F_\beta = (1 + \beta^2) \times \frac{P \times R}{(\beta^2 \times P) + R}
  \label{fbeta}
\end{equation}
\end{minipage}
\vspace{\belowdisplayshortskip}

One issue of using edit overlap to measure performance is that there is often more than one way to define an edit. For example, the edit \edit{has eating}{was eaten} can also be realised as \edit{has}{was} and \edit{eating}{eaten}. If the hypothesis combines them, but the reference does not, the edit will not be counted as a TP even though it produces the same valid correction. As a result, system performance is not measured correctly. 

The innovation of the M$^2$ scorer is that it uses a Levenshtein alignment \citep{levenshtein} between the original text and a system hypothesis to dynamically explore the different ways of combining edits such that the hypothesis edits \textit{maximally match} the reference edits. As such, it overcomes a limitation of the previous scorer used in the HOO shared tasks which could return erroneous scores. Whenever there is more than one set of reference edits for a test sentence, the M$^2$ scorer tries each set in turn and chooses the one that leads to the best performance for that test sentence.  

\subsection{ERRANT}

The ERRANT scorer\footnote{https://github.com/chrisjbryant/errant} \citep{bryant-etal-2017-automatic} is similar to the M$^2$ scorer, in that it is a reference-based metric that measures performance in terms of an edit-based F-score, but differs primarily in that it is also able to calculate error types scores. Specifically, unlike the M$^2$ scorer, it uses a linguistically-enhanced Damerau-Levenshtein alignment algorithm to extract edits from the hypothesis text \citep{felice-etal-2016-automatic}, and then classifies them according to a rule-based error type framework. This facilitates the calculation of F-scores for each error type rather than just overall, which can be invaluable for a detailed system analysis. For example, System A might outperform System B overall, but system B might outperform System A on certain error types, and this information can be used to improve System A.

ERRANT was the first scorer to be able to evaluate GEC systems in terms of error types and is moreover able to do so at three different levels of granularity: 
\begin{itemize}
    \item Edit Operation (3 labels): Missing, Replacement, Unnecessary
    \item Main Type (25 labels): e.g. Noun, Spelling, Verb Tense
    \item Full Type (55 labels): e.g. Missing Noun, Replacement Noun, Unnecessary Noun
\end{itemize}
It is also able to carry out this analysis in terms of both error detection and correction. ERRANT currently only supports English, but other researchers have independently extended it for German \citep{boyd2018using}, Greek \citep{korre-etal-2021-elerrant}, Arabic \citep{belkebir-habash-2021-automatic} and Czech \citep{naplava-2022-czech}.

\subsection{GLEU}

Like M$^2$ and ERRANT, GLEU\footnote{\url{https://github.com/cnap/gec-ranking}} \citep{napoles2015,gleu+} is also a reference-based metric except it does not require explicit edit annotations but rather only corrected reference sentences. It was inspired by the BLEU score \citep{papineni-etal-2002-bleu} commonly used in machine translation and was motivated by the fact that human-annotated edit spans are somewhat arbitrary and time-consuming to collect. The main intuition behind GLEU is that it rewards hypothesis n-grams that overlap with the reference but not the original text, and penalises hypothesis n-grams that overlap with the original text but not the reference. It is important to be aware that GLEU is often attributed to \citet{napoles2015}, but actually implemented according to \citet{gleu+}, which is a revised formulation. The revised formulation is calculated as follows.

Consider a corpus of original sentences $O = \{o_1,...,o_k\}$ and their corresponding hypothesis sentences $H = \{h_1,...,h_k\}$ and reference sentences $R = \{r_1,...,r_k\}$. For each original, hypothesis and reference sentence, let $o_i$, $h_i$ and $r_i$ respectively denote the sequences of n-grams of length $n = \{1, 2, ..., N\}$ ($N=4$ by default in GLEU) in the sentences rather than the sentences themselves. This can then be used to calculate a precision term $p_n$ (\autoref{gleueq}) that takes the intuition about rewarding or penalising n-gram overlap into account. 

\begin{equation}
\begin{gathered}
\resizebox{\textwidth}{!}{$ p_n = \dfrac{\sum\limits_{i=1}^{|H|}\left( \sum\limits_{g \in \{ h_i \cap r_i \}} \textrm{count}_{h_i,r_i}(g) - \sum\limits_{g \in \{ h_i \cap o_i \}} \textrm{max}[0, \textrm{count}_{h_i,o_i}(g) - \textrm{count}_{h_i,r_i}(g)] \right) }{\sum\limits_{i=1}^{|H|} \sum\limits_{g \in \{ h_i \}} \textrm{count}_{h_i}(g) } $} \\
\begin{aligned}
\textrm{count}_{a}(g) &= \textrm{\# occurrences of n-gram }g\textrm{ in }a \\
\textrm{count}_{a,b}(g) &= \textrm{min(\# occurrences of n-gram }g\textrm{ in }a, \textrm{\# occurrences of n-gram }g\textrm{ in }b)
\end{aligned}
\end{gathered}
\label{gleueq}
\end{equation}
\begin{equation}
\textrm{BP} = \begin{cases}
	1 & \textrm{if}\: l_h > l_r \\
	\textrm{exp}(1 - l_r/l_h)  & \textrm{if}\: l_h \leq l_r
\end{cases}
\label{brevpen}
\end{equation}
\begin{equation}
\textrm{GLEU}(O, H, R) = \textrm{BP} \cdot \textrm{exp} \left( \frac{1}{N} \sum\limits_{n=1}^N \textrm{log}~p_n \right)
\label{gleufinal}
\end{equation}

Like the BLEU score, GLEU also has a Brevity Penalty (BP) to penalise hypotheses that are shorter than the references (\autoref{brevpen}), where $l_h$ denotes the total number of tokens in the hypothesis corpus and $l_r$ denotes the total number of tokens in the \textit{sampled} reference corpus. It is important to note that when there is more than one reference sentence, GLEU iteratively selects one at random and averages the score over 500 iterations. GLEU is finally calculated as in \autoref{gleufinal}. 

\subsection{Other Metrics}

In addition to M$^2$, ERRANT and GLEU, other metrics have also been proposed in GEC. Some of these are \textit{reference-based}, i.e. they require human-annotated target sentences, while others are \textit{reference-less}, i.e. they do \textit{not} require human-annotated target sentences. This section briefly introduces metrics of both types.

\subsubsection{Reference-based Metrics}

\paragraph{I-measure} The \textit{I}-measure \citep{felice2015} was designed to overcome certain shortcomings of the M$^2$ scorer, e.g. the M$^2$ scorer is unable to differentiate between a bad system (TP=0, FP$>$0) and a do-nothing system (TP=0, FP=0) which both result in F=0, and instead measure system performance in terms of relative textual \textit{Improvement}. The \textit{I}-measure is calculated by carrying out a 3-way alignment between the original, hypothesis and reference texts and classifying each token according to an extended version of the Writer-Annotator-System (WAS) evaluation scheme \citep{chodorow-etal-2012-problems}. This ultimately enables the calculation of accuracy, which \citet{felice2015} modify to weight TPs and FPs differently to more intuitively reward or punish a system. Having calculated a weighted accuracy score for a system, a baseline weighted accuracy score is computed in the same manner using a copy of the original text as the hypothesis. The difference between these scores is then normalised to fall between -1 and 1, where $I < 0$ indicates text degradation and $I > 0$ indicates text improvement.

\paragraph{GMEG} The GMEG metric \citep{napoles-etal-2019-enabling} is an ensemble metric that was designed to correlate with human judgements on three different datasets. It was motivated by the observation that different metrics correlate very differently with human judgements in different domains, and so a better metric would be more consistent. As an ensemble metric, GMEG depends on features (e.g. precision and recall) from several other metrics, including M$^2$, ERRANT, GLEU, and the \textit{I}-measure (73 features in total). The authors then use these features to train a ridge regression model that was optimised to predict the human scores for different systems. 

\paragraph{GoToScorer} The GoToScorer \citep{gotou-etal-2020-taking} was motivated by the observation that some errors are more difficult to correct than others yet all metrics treat them equally. The GoToScorer hence models error difficulty by weighting edits according to how many different systems were able to correct them; e.g., edits that were successfully corrected by all systems would yield a smaller reward than those successfully corrected by fewer systems. Although this methodology confirmed the intuition that some errors types were easier to correct than others, e.g. spelling errors (easy) vs. synonym errors (hard), one disadvantage of this approach is that the difficulty weights depend entirely on the type and number of systems involved. Consequently, results do not generalise well and error difficulty (or gravity) remains an unsolved problem. 

\paragraph{SE\textsl{\textsc{r}}C\textsl{\textsc{l}}/SERRANT} SE\textsl{\textsc{r}}C\textsl{\textsc{l}} \citep{choshen-etal-2020-classifying} is not a metric \textit{per se}, but rather a method of automatically classifying grammatical errors by their syntactic properties using the Universal Dependencies formalism \citep{nivre-etal-2020-universal}. It is hence similar to ERRANT except it can more easily support other languages. The main disadvantage of SE\textsl{\textsc{r}}C\textsl{\textsc{l}} is that it is not always meaningful to classify errors entirely based on their syntactic properties, e.g. spelling and orthography errors, and some error types are not very informative, e.g. ``VERB$\rightarrow$ADJ''. SERRANT \citep{choshen2021serrant} is hence a compromise that attempts to combine the advantages of both SE\textsl{\textsc{r}}C\textsl{\textsc{l}}  and ERRANT.

\paragraph{PT-M$^2$} The pretraining-based MaxMatch (PT-M$^2$) metric \citep{gong-etal-2022-revisiting} is a hybrid metric that combines traditional edit-based metrics, such as M$^2$, with recent pretraining-based metrics, such as BERTScore \citep{bert-score}. The main advantage of pretraining-based metrics over edit-based metrics is that they are more capable of measuring the semantic similarity between pairs of sentences, rather than just comparing edits. Since \citet{gong-etal-2022-revisiting} found that off-the-shelf pretraining metrics correlated poorly with human judgements on GEC at the sentence level, they instead proposed measuring performance at the edit level. This approach ultimately produced the highest correlation with human judgements on the CoNLL-2014 test set to date, but should be considered with caution, as \citet{hanna-bojar-2021-fine} also highlight some of the limitations of pretraining metrics and cite sources that claim correlation with human judgements may not be the best way to evaluate a metric (see \autoref{sec:reliability}). 

\subsubsection{Reference-less Metrics}

\paragraph{GBMs} The first work to explore the idea of a reference-less metric for GEC \citep{napoles-etal-2016-theres} was inspired by similar work on \textit{quality estimation} in machine translation (e.g. \citet{specia-etal-2020-findings-wmt}). Specifically, the authors proposed three Grammaticality-Based Metrics (GBMs) that either use a benchmark GEC system to count the errors in the output produced by other GEC systems or else predict a grammaticality score using a pretrained ridge regression model \citep{Heilman-et-al-2014}. The main limitation of these metrics is that they i) require an existing GEC system to evaluate other GEC systems and ii) are insensitive to changes in meaning. The authors thus proposed interpolating reference-less metrics with other reference-based metrics.

\paragraph{GFM} \citet{asano-etal-2017-reference} extended the work on GBMs by introducing three reference-less metrics for Grammaticality, Fluency and Meaning preservation (GFM). Specifically, the Grammaticality metric combines \citeauthor{napoles-etal-2016-theres}'s \citeyear{napoles-etal-2016-theres} GBMs into a single model, the Fluency metric computes a score using a language model, and the Meaning preservation metric computes a score using the METEOR metric from machine translation \citep{denkowski-lavie-2014-meteor}. A weighted linear sum of the three scores is then used as the final score. The main weaknesses of the GFM metric are that the Grammaticality and Fluency metrics suffer from the same limitations as GBMs, and the Meaning preservation metric only models shallow text similarity in terms of overlapping content words.

\paragraph{US{\textsl{\textsc{im}}}} The US\textsc{im} metric \citep{choshen-abend-2018-reference} was motivated by the fact that no other metric takes deep semantic similarity into account and it is possible that a GEC system might change the intended meaning of the original text; e.g., by inserting/deleting `not' or replacing a content word with an incorrect synonym. It is calculated by first automatically annotating the original and hypothesis texts as semantic graphs using the UCCA semantic scheme \citep{abend-rappoport-2013-universal} and then measuring the overlap between the graphs (in terms of matching edges) as an F-score. US\textsc{im} was thus designed to operate as a complementary metric to other metrics. 

\paragraph{SOME} Sub-metrics Optimised for Manual Evaluation (SOME) \citep{yoshimura2020-reference} is an extension of GFM that was designed to optimise each Grammaticality, Fluency and Meaning preservation metric to more closely correlate with human judgements. The authors achieved this by annotating the system output of five recent systems on a 5-point scale for each metric and then fine-tuning BERT \citep{devlin-etal-2019-bert} to predict these human scores. This differs from GFM in that GFM was fine-tuned to predict the human ranking of different systems rather than explicit human scores. While the authors found SOME correlates more strongly with human judgements than GFM, both metrics nevertheless suffer from the same limitations. 

\paragraph{Scribendi Score} The Scribendi Score \citep{islam-magnani-2021-end} was designed to be simpler than other reference-less metrics in that it requires neither an existing GEC system nor fine-tuning. Instead, it calculates an absolute score (1=positive, -1=negative, 0=no change) from a combination of language model perplexity (GPT2: \citet{radford2019language}) and sorted token/Levenshtein distance ratios, which respectively ensure that i) the corrected sentence is more probable than the original and ii) both sentences are not significantly different from each other. While it is intuitive that these scores correlate with the grammaticality of a sentence, they are not, however, a robust way of evaluating a GEC system. For example, the sentence `I saw \textit{the} cat'' is more probable than ``I saw \textit{a} cat'' in GPT2 (160.8 vs 156.4), and both sentences are moreover very similar, yet we would not want to always reward this as a valid correction since both sentences are grammatical. We observe the same effect in ``I ate the cake.'' (130.2) vs. ``I ate the pie.'' (230.7) and so conclude that the Scribendi Score is highly likely to erroneously reward false positives. 

\paragraph{IMPARA} The Impact-based Metric for GEC using Parallel data (IMPARA) \citep{maeda-etal-2022-impara} is a hybrid reference-based/reference-less metric that requires parallel data to train an edit-based quality estimation and semantic similarity model, but can be used as a reference-less metric after training. It is sensitive to the corpus it is trained on (i.e., it does not generalise well to unseen domains) but shows comparable or better performance to SOME in terms of correlation with human judgements. Its main advantage is that it only requires parallel data for training (i.e., not system output or human judgements), but its main disadvantage is that IMPARA scores are not currently interpretable by humans.

\subsection{Metric Reliability}
\label{sec:reliability}

Given the number of metrics that have been proposed, it is natural to wonder which metric is best. This is not straightforward to answer, however, as all metrics have different strengths and weaknesses. There has nevertheless been a great deal of work based on the assumption that the ``best'' metric is the one that correlates most closely with ground-truth human judgements.  

With this in mind, the first work to compare metric performance with human judgements was by \citet{napoles2015} and \citet{grundkiewicz2015}, who independently collected human ratings for the 13 system outputs from the CoNLL-2014 shared task (including the unchanged original text) using the Appraise evaluation framework \citep{federmann-2010-appraise} commonly used in MT. This framework essentially asks humans to rank randomly chosen samples of 5 system outputs (ties are permitted) in order to build up a collection of pairwise judgements that can be used to extrapolate an overall system ranking. A metric can then be judged in terms of how well it correlates with this extrapolated ranking. The judgements collected by \citet{grundkiewicz2015} in particular proved especially influential (their dataset was much larger than \citet{napoles2015}) and were variously used to justify GLEU as a better metric than M$^2$ \citep{napoles2015,napoles-etal-2016-theres,sakaguchi-etal-2016-reassessing} and motivate almost all reference-less metrics to date (except US\textsc{im}). 

Unfortunately however, this methodology was later found to be problematic and many of the conclusions drawn using these datasets were thrown into doubt. Notable observations included:
\begin{itemize}
    \item The correlation coefficients reported by \citet{napoles2015} and \citet{grundkiewicz2015} were very different even though they essentially carried out the same experiment (albeit on different samples) \citep{choshen2018}.
    \item This method of human evaluation was abandoned in machine translation due to unreliability \citep{choshen2018, graham-etal-2015-accurate}.
    \item \citet{chollampatt-ng-2018-reassessment} found no evidence of GLEU being a better metric than M$^2$ for ranking systems.
\end{itemize}

\citet{choshen2018} surmise that one of the reasons these metric correlation experiments proved unreliable is that rating sentences for grammaticality is a highly subjective task which often shows very low inter-annotator agreement (IAA); e.g. it is difficult to determine whether a sentence containing one major error should be considered ``more grammatical'' than a sentence containing two minor errors. 

\citet{napoles-etal-2019-enabling} nevertheless carried out a follow-up study which not only used a continuous scale to judge sentences (rather than rank them) \citep{sakaguchi-van-durme-2018-efficient}, but also collected judgements on all pairs of sentences to overcome sampling bias. They furthermore reported results on different datasets from different domains, rather than just CoNLL-2014, in an effort to determine the most generalisable metric. Their results, partially recreated in \autoref{tab:metric_correlation}, hence found that dataset does indeed have an effect on metric performance, most likely because different error type distributions are judged inconsistently by humans. In fact, although \citet{napoles-etal-2019-enabling} reported very high IAA at the corpus level (0.9-0.99 Pearson/Spearman), IAA at the sentence level was still low to average (0.3-0.6 Pearson/Spearman). 

\begin{table}[th!]
  \centering
    \begin{tabular}{l|r|r|r|r|r|r}
          & \multicolumn{2}{c|}{FCE} & \multicolumn{2}{c|}{Wiki} & \multicolumn{2}{c}{Yahoo} \\
    Metric & $r$ & $\rho$ & $r$ & $\rho$ & $r$ & $\rho$ \\
    \hline
    ERRANT F$_{0.5}$ & \textbf{0.919} & \textbf{0.887} & 0.401 & 0.555 & 0.532 & 0.601 \\
    GLEU  & 0.838 & 0.813 & 0.426 & 0.538 & 0.740 & 0.775 \\
    \textit{I}-measure & 0.819 & 0.839 & \textbf{0.854} & \textbf{0.875} & \textbf{0.915} & \textbf{0.900} \\
    M$^2$ F$_{0.5}$ & 0.860 & 0.849 & 0.346 & 0.552 & 0.580 & 0.699 \\
    \end{tabular}
  \caption{Pearson $r$ and Spearman $\rho$ correlation coefficients for different metrics across three different datasets. This is a subset of the results reported in \citet{napoles-etal-2019-enabling} Table 8. }
  \label{tab:metric_correlation}%
\end{table}%

Ultimately, although ground-truth human judgements may be an intuitive way to benchmark metric performance, they are also highly subjective and should be considered with caution. Nothing demonstrates this sentiment better than the conclusions drawn about the \textit{I}-measure, which was initially found to have a weak negative correlation with human judgements \citep{napoles2015,grundkiewicz2015,sakaguchi-etal-2016-reassessing}, subsequently found to have good correlation at the sentence level \citep{napoles-etal-2016-theres} and finally considered the best singular metric across multiple domains \citep{napoles-etal-2019-enabling}. Reliable methods of evaluating automatic metrics thus remain an active area of research. 

\subsection{Evaluation Best Practices}

A common pitfall for new researchers in GEC concerns which metric to use with which dataset; for example the M\textsuperscript{2} scorer with JFLEG, or the \textit{I}-measure with BEA-2019. While there is no empirical reason to prefer one metric over another, in practice, the most popular GEC test sets are almost always evaluated with a single, specific metric:

\begin{itemize}
    \item CoNLL-2014 is evaluated with the M\textsuperscript{2} scorer
    \item JFLEG is evaluated with GLEU
    \item BEA-2019 is evaluated with ERRANT
\end{itemize}

This choice of experimental setup is largely motivated by historical reasons (e.g., GLEU and ERRANT did not exist during CoNLL-2014), but has nevertheless persisted in order to ensure fair comparison with subsequent work. One particularly common mistake is to evaluate CoNLL-2014 with ERRANT or BEA-2019 with the M\textsuperscript{2} scorer because both metrics return an F-score, yet M\textsuperscript{2} F\textsubscript{0.5} is not equivalent to ERRANT F\textsubscript{0.5} \citep{bryant-etal-2017-automatic}. It is thus imperative that a dataset be evaluated with its associated metric in order to facilitate a meaningful comparison.

\subsubsection{Caveats} \hfill \\

\noindent Despite this convention, it is also important to highlight the limitations of this setup, as it is not always desirable to optimise different systems for different test sets using different metrics. Instead, we should remember that the ultimate goal of GEC is to build systems that generalise well, and so we should not place too much emphasis on specific experimental configurations. It is with this in mind that \citet{mita-etal-2019-cross} recommend evaluating on multiple corpora in order to reveal any systematic biases towards particular domains or user demographics, while \citet{napoles-etal-2019-enabling} recommend evaluating using their trained metric that was designed to be less sensitive to dataset biases. These approaches thus add greater confidence that a model is versatile and does not overfit to a specific type of input.

\subsubsection{Recommendations} \hfill \\

\noindent In light of the confusion surrounding different experimental setups, we make the following recommendations for ensuring a meaningful comparison in English GEC evaluation. This is not an exhaustive list, but we attempt to summarise the current standard experimental setups that facilitate the most informative comparison with previous work.

\begin{enumerate}
    \item \textit{Evaluate on the BEA-2019 test set using ERRANT.} \\
    The BEA-2019 test set is one of the most diverse test sets that contains texts from the full range of learner backgrounds and ability levels on a wide range of topics. This makes it a good benchmark for system robustness and generalisability. It is also the official test set of the most recent shared task.
    
    \item \textit{Evaluate on the CoNLL-2014 test set using the M\textsuperscript{2} scorer.} \\
    The CoNLL-2014 test set is one of the most well-known test sets that has been widely used to benchmark progress in the field; it is thus an important indicator of system performance. It is also the official test set of the second most recent shared task.

    \item \textit{Evaluate on the GMEG and/or CWEB test sets using ERRANT. }\\
    One of the main limitations of the BEA-2019 and CoNLL-2014 test sets is that they mainly represent non-native language learners. It can therefore be beneficial to evaluate on native speaker errors in GMEG and CWEB to obtain a more complete picture of system generalisability. 
 
    \item \textit{Evaluate on JFLEG using GLEU.} \\
    The main reason to evaluate on JFLEG is to test systems on more complex fluency edits rather than minimal edits. Not all edits in JFLEG are fluency edits however, and the test set is very small, so researchers have seldom reported GLEU on JFLEG in recent years \citep{gong-etal-2022-revisiting}.
\end{enumerate}

Ultimately, robust evaluation is rarely as straightforward as directly comparing one number against another, and it is important to consider, for example, whether a model has been trained/fine-tuned on in-domain data, optimised for a specific metric, or only evaluated on a specific target test set. Each of these factors impacts how a score should be interpreted, especially in relation to previous work, and there is a real danger of rewarding a highly-optimised, specialised system, over a lower-scoring but more versatile system that may actually be more desirable.

\section{System Comparison}
\label{sec:sys_comp}

In this section, we compare the most recent state-of-the-art systems from the past couple of years and comment on the innovations that led to them performing better than previous work. The full list of systems we compare is shown in \autoref{tab:sota}. For a comparison of systems between 2014-2020, we refer the reader to \citet[Table 7]{wang2020comprehensive}.

\subsection{System Description}

We first note that many of the systems in \autoref{tab:sota} are extensions of 3 other systems: \citet{omelianchuk-etal-2020-gector}, \citet{sun-etal-2021-instantaneous}, and \citet{kiyono-etal-2019-empirical}. Specifically, \citet{omelianchuk-etal-2020-gector} built a sequence tagging model (\autoref{sec:seq2edits}) using a pre-trained language model (e.g. BERT) and 9 million synthetic sentence pairs, \citet{sun-etal-2021-instantaneous} used a rule-based approach to generate 300 million synthetic sentence pairs (\autoref{sec:noise-injection}) to train a modified BART model which contains 12 encoders and 2 decoders, and \citet{kiyono-etal-2019-empirical} used 70 million synthetic sentence pairs generated through back-translation (\autoref{sec:back-translation}) to train a Transformer-big model. 

Many of these systems specifically build on top of \citet{omelianchuk-etal-2020-gector}, including systems from \citet{sorokin-2022-improved, lai-etal-2022-type, parnow-etal-2021-grammatical, yasunaga-etal-2021-lm}. Specifically, \citet{sorokin-2022-improved} and \citet{tarnavskyi-etal-2022-ensembling} upgraded the pre-trained language model from base to large (e.g., RoBERTa-base vs. RoBERTa-large) and employed an additional mechanism to select the final edits by means of edit-scoring or majority voting (VT) respectively. \citet{parnow-etal-2021-grammatical} and \citet{lai-etal-2022-type} address the problem of edit interdependence, i.e. when the correction of one error depends on another, by means of GANs and multi-turn training respectively. \citet{yasunaga-etal-2021-lm} applied the break-it-fix-it (BIFI) framework \citep{yasunaga2021break} to \citet{omelianchuk-etal-2020-gector} (\autoref{sec:lm}) to gradually train a system that iteratively generates and learns from more realistic synthetic data. In contrast, \citet{sun-wang-2022-adjusting} add a single hyperparameter to \citet{sun-etal-2021-instantaneous} to control the trade-off between precision and recall (PRT), \citet{kaneko-etal-2020-encoder} incorporate BERT into \citet{kiyono-etal-2019-empirical} (\autoref{sec:transformer}), and \citet{mita-etal-2020-self} applied a self-refinement data augmentation strategy to \citet{kiyono-etal-2019-empirical} (\autoref{sec:aug_off}). 

\begin{landscape}
\begin{table}[p]
  \centering
  \footnotesize
  \setlength\tabcolsep{5pt}
\begin{threeparttable}
    \begin{tabular}{L{5cm}|R{1.5cm}|L{1.65cm}|L{1.8cm}|L{2.3cm}|l|L{1.6cm}|L{1.6cm}}
    \hline
    \textbf{System} & \textbf{Synthetic Sents} & \textbf{Corpora} & \textbf{Pre-trained Model} & \textbf{Architecture} & \textbf{Techniques} & \textbf{CoNLL14 M2} & \textbf{BEA19 ERRANT} \\
    \hline
    \citet{qorib22}   & - & W (dev) & Various\tnote{1} & T5-large, RoBERTa-base, XLNet-base, Transformer-big & SC    & 69.5  & 79.9 \\
    \hline
    \citet{lai-etal-2022-type} & 9m    & N+F+L+W & RoBERTa, XLNet & RoBERTa-base, XLNet-base & ENS+PRT+MTD & 67.0  & 77.9 \\
    \hline
    \citet{sorokin-2022-improved} & 9m & cL+N+F+W & RoBERTa & RoBERTa-large & RE+MTD & 64.0 & 77.1 \\
    \hline
    \citet{tarnavskyi-etal-2022-ensembling} & - & N+F+L+W & RoBERTa, XLNet, DeBERTa & RoBERTa-large, XLNet-large, DeBERTa-large & VT+PRT+MTD & 65.3  & 76.1 \\
    \hline
    \citet{rothe-etal-2021-simple} & - & cL    & T5-xxl & T5-xxl & - & 68.9 & 75.9 \\
    \hline
    \citet{sun-wang-2022-adjusting} & 300m  & N+F+L+W & BART  & BART (12+2) & PRT   & - & 75.0 \\
    \hline
    \citet{stahlberg-kumar-2021-synthetic}    & 546m  & F+L+W & - & Transformer-big & ENS   & 68.3  & 74.9 \\
    \hline
    \citet{cao-etal-2022-eb}    & 200m  & cL+N+F+W & - & Transformer-big & ENS   & 68.5  & 74.8 \\
    \hline
    \citet{omelianchuk-etal-2020-gector}    & 9m    & N+F+L+W & BERT, RoBERTa, XLNet & BERT-base, RoBERTa-base, XLNet-base & ENS+PRT+MTD & 66.5  & 73.7 \\
    \hline
    \citet{lichtarge-etal-2020-data}    & 340m  & F+L+W & - & Transformer-big & ENS   & 66.8  & 73.0 \\
    \hline
    \citet{zhang-etal-2022-syngec} & - & cL+N+F+W  & BART & BART-large & - & 67.6 & 72.9 \\  
    \hline
    \citet{sun-etal-2021-instantaneous} & 300m  & N+F+L+W & BART  & BART (12+2) & - & 66.4  & 72.9 \\
    \hline
    \citet{yasunaga-etal-2021-lm} & 9m    & N+F+L+W & XLNet & XLNet-base & PRT+MTD & 65.8  & 72.9 \\
    \hline
    \citet{parnow-etal-2021-grammatical}    & 9m    & N+F+L+W & XLNet & XLNet-base & PRT+MTD & 65.7  & 72.8 \\
    \hline
    \citet{yuan-etal-2021-multi} & - & N+F+L+W +CLC & ELECTRA & Multi-encoder, Transformer-base & RE    & 63.5  & 70.6 \\
    \hline
    \citet{stahlberg-kumar-2020-seq2edits}    & 346m  & F+L+W & - & Seq2Edits (modified Transformer-big) & ENS+RE & 62.7  & 70.5 \\
    \hline
    \citet{kaneko-etal-2020-encoder}    & 70m   & N+F+L+W & - & Transformer-big & ENS+RE & 65.2  & 69.8 \\
    \hline
    \citet{mita-etal-2020-self}    & 70m   & N+F+L+W & - & Transformer-big & ENS+RE & 63.1  & 67.8 \\
    \hline
    \citet{chen-etal-2020-improving-efficiency}    & 260m  & N+F+L+W & RoBERTa & Transformer-big & - & 61.0  & 66.9 \\
    \hline
    \citet{katsumata-komachi-2020-stronger}    & - & N+F+L+W & BART  & BART-large & ENS   & 63.0  & 66.1 \\
    \hline
\end{tabular}%
\begin{tablenotes}\footnotesize
\item[1] Combines \citet{rothe-etal-2021-simple,omelianchuk-etal-2020-gector,kiyono-etal-2019-empirical,grundkiewicz-etal-2019-neural,choe-etal-2019-neural}.  
\end{tablenotes}
\end{threeparttable}
\caption{Top-performing systems since 2020. The symbols in the Corpora column are N: NUCLE, F: FCE, L: Lang-8, W: W\&I, cL: cLang-8, and CLC: Cambridge Learner Corpus. The symbols in the Techniques column are ENS: ensemble, MTD: multi-turn decoding, PRT: precision-recall trade-off, RE: re-ranking, SC: system combination, and VT: voting combination.}
\label{tab:sota}
\end{table}%
\end{landscape}

Other systems include \citet{katsumata-komachi-2020-stronger} and \citet{rothe-etal-2021-simple}, who respectively explored the effectiveness of using pre-trained BART \citep{lewis-etal-2020-bart} and T5 \citep{2020t5} as the base model for GEC; \citet{zhang-etal-2022-syngec} subsequently extended \citet{katsumata-komachi-2020-stronger} by adding syntactic information (\autoref{sec:transformer}). \citet{chen-etal-2020-improving-efficiency} and \citet{yuan-etal-2021-multi} meanwhile both combined error detection with error correction by respectively constraining the output of a GEC system based on a separate GED system and jointly training GED as an auxiliary task (\autoref{sec:multi-learn}). \citet{stahlberg-kumar-2020-seq2edits} proposed a seq2edit approach that explicitly predicts a sequence of tuple edit operations to apply to an input sentence (\autoref{sec:seq2edits}), while \citet{stahlberg-kumar-2021-synthetic} developed a method to generate a specific type of error in a sentence (given a clean sentence and an error tag), which could be used to generate synthetic datasets that more closely match the error distribution in a real corpus (\autoref{sec:back-translation}). Finally, \citet{lichtarge-etal-2020-data} used delta-log-perplexity to weight the contribution of each sentence in the training set towards overall model performance, downweighting those that added the most noise (\autoref{sec:aug_off}), and \citet{qorib22} used a binary classifier based on logistic regression to combine multiple GEC systems using only the output from each individual component system. 

\subsection{Analysis}

Despite all these enhancements, we first observe that it is very difficult to draw conclusions about the efficacy of different techniques in \autoref{tab:sota}, because different systems were trained using different amounts/types of data (both real and artificial) and developed using different pre-trained models and performance-boosting techniques. Consequently, the systems are rarely directly comparable and we can only infer the relative advantages of different approaches from the wider context. With this in mind, the general trend in the past couple of years has been to scale models up using i) more artificial data, ii) multiple pre-trained models/architectures, and iii) multiple performance-boosting techniques.

In terms of artificial data, the trend is somewhat mixed, as on the one hand, \citet{stahlberg-kumar-2021-synthetic} introduced a system trained on more than half a billion synthetic sentences, but on the other hand, they were still outperformed by systems that used orders of magnitude less data \citep{lai-etal-2022-type,tarnavskyi-etal-2022-ensembling}. This pattern has been consistent for several years now and reveals a delicate trade-off between artificial data quantity and quality. There is ultimately no clear relationship between data quantity and performance, and some systems still achieve competitive performance without artificial data \citep{rothe-etal-2021-simple,yuan-etal-2021-multi,katsumata-komachi-2020-stronger}.

The use of several pre-trained model architectures, however, tells a different story and it is generally the case that using multiple architectures improves performance: the top 3 latest state-of-the-art systems all use at least 2 different pre-trained models \citep{qorib22,lai-etal-2022-type,tarnavskyi-etal-2022-ensembling}. This suggests that different pre-training tasks capture different aspects of natural language that complement each other in different ways in GEC. In contrast, approaches that rely on a single pre-trained model typically perform slightly worse than those that combine architectures, although it is worth keeping in mind that there is also a trade-off between model complexity and run-time which is seldom reported \citep{omelianchuk-etal-2020-gector,sun-etal-2021-instantaneous}.

Finally, adding more performance-boosting techniques also tends to result in better performance, and the systems that incorporate the most techniques typically score highest. Among these techniques, the use of model ensembling or system combination (\autoref{sec:sys_comb}) mitigates the instability of neural models and allows a final system to make use of the strengths of several other systems. However, this comes at a cost to model complexity and run-time.

\section{Future Challenges}
\label{sec:chal}

While much progress has been made in the past decade, several important challenges remain \citep{qorib-ng-2022-grammatical}. This section highlights some of them and offers suggestions for future work.

\paragraph{Domain Generalisation} Robustness is an important attribute of any NLP system. In the case of GEC, we not only want our systems to work well for language learners, but also native speakers in different domains such as business emails, literature and instruction manuals. Some efforts have been made in this direction, such as the native web texts in CWEB \citep{flachs-etal-2020-grammatical}, scientific articles in AESW \citep{Daudaravicius-et-al-2016} and conversational dialog in ErAConD \citep{yuan-etal-2022-eracond}, but more effort is needed to create new corpora that represent a wider variety of domains. This is important because previous research has shown that systems that perform well in one domain do not necessarily perform well in other domains \citep{napoles-etal-2019-enabling}.

\paragraph{Personalised Systems} Related to domain generalisation is the fact that system performance is also tied to the profiles of the users in the training data. For example, a system trained on L2 English data produced by advanced L1 Spanish learners is unlikely to perform as well on L2 English data produced by beginner L1 Japanese learners because of the mismatch in ability level and first language. It is thus important to develop corpora and tools that can adapt to different users \citep{chollampatt-etal-2016-adapting}, since different ability levels and L1s can significantly affect the distribution of errors that authors are likely to make \citep{nadejde-tetreault-2019-personalizing}.

\paragraph{Feedback Comment Generation} GEC systems are currently trained to correct errors without explaining why a correction was needed. This is insufficient in an educational context however, where it is desirable for a system to explain the cause of an error such that a user may learn from it and not make the same mistake again. Resources have begun to emerge to support this endeavour, but much more work is needed to generate robust feedback comments to support explainable GEC \citep{nagata-2019-toward,nagata-etal-2020-creating,hanawa-etal-2021-exploring,nagata-etal-2021-shared}.

\paragraph{Model Interpretability} Related to feedback generation, it is also important that model output is interpretable by humans. For example, although a system may make a prediction with high confidence, there is no guarantee that the prediction will be consistent with human intuition. Researchers have thus begun to build systems that estimate the quality of model output in an effort to provide more confidence that a given prediction is correct \citep{chollampatt-ng-2018-neural,liu-etal-2021-neural}. Similarly, \citet{kaneko-etal-2022-interpretability} propose an example-based approach, where a model additionally outputs similar corrections in different contexts in order to add credibility to the notion that the model truly understood the error. 

\paragraph{Semantic Errors} One of the areas where state-of-the-art systems still underperform is semantic errors, which include complex phenomenon such as collocations, idioms, multi-word expressions and fluency edits. A lot of work in GEC has focused on correcting function word errors, which typically have small confusion sets and comprise a majority of error types, but this does not mean we can neglect the correction of content word errors. Although there has been some work on correcting collocations \citep{kochmar-briscoe-2014-detecting,herbelot-kochmar-2016-calling} and multi-word expressions \citep{mizumoto-etal-2015-grammatical,taslimipoor-etal-2022-improving}, semantic errors remain a notable area in which GEC systems could improve. 

\paragraph{Contextual GEC} To date, most GEC systems operate at the sentence level, and so do not perform well on errors that require cross-sentence context or document-level understanding. Although work has already been done to incorporate multi-sentence context into GEC systems \citep{chollampatt-etal-2019-cross,yuan-bryant-2021-document,mita2022automated}, almost all current datasets expect sentence tokenised input and so do not facilitate multi-sentence evaluation. Paragraph or document-level datasets, like in the Arabic QALB shared tasks \citep{mohit-etal-2014-first,rozovskaya-etal-2015-second}, should thus be developed to encourage contextual GEC in the future. 

\paragraph{System Combination} Although much recent work focuses on NMT for GEC, this does not mean that other approaches have nothing to offer. Work on system combination has shown that systems built with different approaches have complementary strengths and weaknesses such that a combined system can achieve significantly improved performance \citep{susanto-etal-2014-system,ddc,gec-ip,qorib22}. Better understanding of these strengths and weaknesses, and when and how to combine approaches are promising areas of research. One tool is ALLECS \citep{allecs23}, which is a web-interface tool to produce text corrections using GEC system combination methods.

\paragraph{Training Data Selection} Current state-of-the-art systems rely on pre-training on a massive amount of synthetic parallel corpora, however this is both computationally-expensive and not environmentally friendly. It is also questionable whether so much training data is really necessary, as humans are not exposed to training data on such a massive scale, yet can still correct errors without issue. A more economical approach to effective training data selection is thus an important research question that will go a long way towards reducing training time and developing more efficient GEC systems \citep{lichtarge-etal-2020-data,takahashi-etal-2020-grammatical,mita-yanaka-2021-grammatical,rothe-etal-2021-simple}.

\paragraph{Unsupervised Approaches} The dependency on parallel corpora (both real and synthetic) is a major limiting factor in GEC system development, in that it is both laborious and time-consuming to train human annotators to manually correct errors, and also surprisingly difficult to generate high-quality synthetic errors that reliably imitate human error patterns. It is furthermore noteworthy that humans can correct errors without access to a large corpus of erroneous examples and instead rely on their knowledge of grammatical language in order to detect and correct mistakes. It should thus be intuitive that a GEC system might be able to do the same by interpreting deviations from grammatical data as anomalies that need to be corrected. The success of such an unsupervised approach would significantly hasten the development of multilingual GEC systems and also eliminate the need to compile parallel corpora.

\paragraph{Multilingual GEC} Although most work on GEC has focused on English, work on other languages is also beginning to take off as new resources become available; e.g. in German \citep{boyd2018using}, Russian \citep{rozovskaya-roth-2019-grammar} and Czech \citep{naplava-2022-czech}. While it is important to encourage research into GEC systems for specific languages, it is also important to remember that it is ultimately not scalable to build a separate system for every language. It is desirable to work towards a single multilingual system that can correct all languages simultaneously like in machine translation \citep{katsumata-komachi-2020-stronger,rothe-etal-2021-simple}.

\paragraph{Spoken GEC} Another aspect of GEC that has seldom been explored in the literature is that of spoken GEC. While progress has largely been hindered by a lack of available data, researchers have recently begun to build systems capable of detecting and correcting errors in learner speech \citep{knill-etal-2019-automatic,caines-etal-2020-grammatical,KyriakopoulosKG20,LuG020,lu-etal-2022-assessing}. Compared to text-based GEC, additional challenges include recognising non-native accented speech (possibly including non-standard pronunciation), disfluency detection, and utterance segmentation.

\paragraph{Improved Evaluation} Finally, robust evaluation of GEC system output is still an unsolved problem and current evaluation practices may actually hinder progress \citep{rozovskaya-roth-2021-good}. For example, almost all metrics to date require tokenised text, yet end-users require untokenised text, and so there is a disconnect between system capability and user expectation. Similarly, GEC systems are typically trained to output a single best correction for a sentence, yet end-users may prefer a short n-best list of possible corrections for each edit, like in most spellcheckers. Ultimately, alternative answers and untokenised text are not yet properly accounted for in GEC system evaluation, leaving room for new metrics to drive the field towards better practices. 

\section{Conclusion}
\label{sec:conc}

In this survey paper, we set out to provide a comprehensive overview of the state of the art in the field of Grammatical Error Correction. Our main goal was to summarise the progress that has been made since \citet{leacock2014automated} but also complement the work of \citet{wang2020comprehensive} with more in-depth and recent coverage on various topics. 

With this in mind, we first explored the nature of the task and illustrated the inherent difficulties in defining an error according to the perceived communicative intent of the author. We next alluded to how these difficulties can manifest in human-annotated corpora, before introducing the most commonly used benchmark corpora for English, several less commonly used corpora for English, and new corpora for GEC systems in other languages, including Arabic, Chinese, Czech, German and Russian. Research into GEC for non-English languages has begun to take off in the last couple of years and will no doubt continue to grow in the future. 

We next characterised the evolution of approaches to GEC, from error-type specific classifiers to state-of-the-art NMT and edit-based sequence-labelling, and summarised some of the additional supplementary techniques that are commonly used to boost performance, such as re-ranking, multi-task learning and iterative decoding. We also described different methods of artificial data generation and augmentation, which have become core components of recent GEC systems, but also drew attention to the benefits of low-resource GEC systems that may be less resource intensive and more easily extended to other languages. 

Robust evaluation is still an unsolved problem in GEC, but we introduced the most commonly used metrics in the field, along with their strengths and weaknesses, and listed previous attempts at both reference-based and reference-less metrics that were designed to overcome various shortcomings. We furthermore highlighted the difficulty in correlating human judgements with metric performance in light of the highly subjective nature of the task. 

Finally, we provided an analysis of very recent progress in the field, including making observations about which techniques/resources seemed to perform best (particularly in the context of model efficiency), before concluding with several possibilities for future work. We hope that this survey will serve as comprehensive resource for researchers who are new to the field or who want to be kept apprised of recent developments.

\begin{acknowledgments}

We thank the anonymous reviewers for their helpful comments. This research is supported by both Cambridge University Press \& Assessment and the National Research Foundation, Singapore under its AI Singapore Programme (AISG Award No: AISG-RP-2019-014). 

\end{acknowledgments}

\starttwocolumn
\bibliography{compling_style}
\bibliographystyle{compling}

\end{document}